\def\year{2018}\relax
\documentclass[letterpaper]{article} 
\usepackage{aaai18}  
\usepackage{times}  
\usepackage{helvet}  
\usepackage{courier}  
\usepackage{url}  
\usepackage{graphicx}  
\title{On the ERM Principle with Networked Data}

\author{Yuanhong Wang\textsuperscript{1,2}, Yuyi Wang\textsuperscript{3}, Xingwu Liu\textsuperscript{4,5}, Juhua Pu\textsuperscript{1,2}\thanks{Corresponding author}\\
\textsuperscript{1} State Key Laboratory of Software Development Environment, Beihang University, Beijing, China\\
\textsuperscript{2} Research Institute of Beihang University in Shenzhen, Shenzhen, China\\
\textsuperscript{3} Disco Group, ETH Zurich, Switzerland\\
\textsuperscript{4} Institute of Computing Technology, Chinese Academy of Sciences, Beijing, China\\
\textsuperscript{5} University of Chinese Academy of Sciences, Beijing, China
}

\usepackage[utf8]{inputenc} 
\usepackage{booktabs}       
\usepackage{amsfonts}       
\usepackage{microtype}      
\usepackage{algorithm}
\usepackage[noend]{algpseudocode}

\usepackage{ifthen}

\def\GenerateShortVersion{}
\ifthenelse{\isundefined{\GenerateShortVersion}}
{
\def\LongVersion{}
\def\LongVersionEnd{}
\long\def\ShortVersion#1\ShortVersionEnd{}
}
{
\def\ShortVersion{}
\def\ShortVersionEnd{}
\long\def\LongVersion#1\LongVersionEnd{}
}

\ifthenelse{\isundefined{\GenerateSingleColumn}}
{
\def\DoubleColumn{}
\def\DoubleColumnEnd{}
\long\def\SingleColumn#1\SingleColumnEnd{}
}
{
\def\SingleColumn{}
\def\SingleColumnEnd{}
\long\def\DoubleColumn#1\DoubleColumnEnd{}
}

\usepackage{amsmath}
\usepackage{amsthm}
\usepackage{amsfonts}
\usepackage{url}
\usepackage{graphicx}
\usepackage{color,colortbl}
\usepackage{epstopdf}
\definecolor{Gray}{gray}{0.7}
\usepackage{dsfont}
\usepackage{breqn}
\usepackage{mathrsfs}
\usepackage{breakcites}
\usepackage{subfigure}
\usepackage{bm}
\usepackage{multirow}
\usepackage{array,booktabs}
\newcolumntype{P}[1]{>{\centering\arraybackslash}p{#1}}
\newcolumntype{M}[1]{>{\centering\arraybackslash}m{#1}}
\allowdisplaybreaks

\newtheorem{theorem}{Theorem}
\newtheorem{lemma}{Lemma}

\newtheorem{corollary}{Corollary}
\newtheorem{definition}{Definition}
\newtheorem{example}{Example}
\newtheorem*{example*}{Example}

\newtheorem*{remark*}{Remark}
\newtheorem{assumption}{Assumption}
\newtheorem*{assumption*}{Assumption}
\newcommand{\E}{\mathbb{E}}
\newcommand{\Pro}{P}
\newcommand{\Var}{\text{Var}}
\newcommand{\indicator}{\mathds 1}
\newcommand{\overbar}[1]{\mkern 1.5mu\overline{\mkern-1.5mu#1\mkern-1.5mu}\mkern 1.5mu}
\newcommand*\cen[1]{\overbar{#1}}
\newcommand{\weight}{\mathbf{w}}
\newcommand{\probdistri}{\mathbf{p}}
\newcommand{\verticeweight}{\bar{\mathbf{w}}}
\newcommand{\real}{\mathbb{R}}
\newcommand{\xspace}{\mathcal{X}}
\newcommand{\yspace}{\mathcal{Y}}
\newcommand{\zspace}{\mathcal{Z}}
\newcommand{\bayeserror}{L^*}
\newcommand{\empiricalrisk}[1]{L_{#1}}
\newcommand{\distribution}{P}
\newcommand{\risk}{L}
\newcommand{\lossf}{\ell}
\newcommand{\relossf}{\bar{\ell}}
\newcommand{\cenprocess}[1]{\cen{\gamma_{#1}}}
\newcommand{\rademacher}{\sigma}
\newcommand{\edge}[1]{\{#1\}}
\newcommand{\pair}[1]{(#1)}
\newcommand{\lnorm}{\mathbb{L}}
\newcommand{\lnormspace}{\mathscr{L}}

\newcommand{\problemabbr}{\textnormal{C}\textsc{lanet}}
\newcommand{\normo}[1]{\|#1\|_1}
\newcommand{\fcoloring}{\chi^*}
\newcommand{\complexbound}{\beta}
\newcommand{\set}[1]{\{#1\}}
\newcommand{\trainingset}{\mathbb{S}}
\newcommand{\rademachercomplexity}{\mathfrak{R}}

\newcommand{\citet}[1]{\citeauthor{#1}\ (\citeyear{#1})}
\newcommand{\citett}[2]{\citeauthor{#2}\ (\citeyear{#2}, #1)}
\newcommand{\citep}[3]{(#1\ \citeauthor{#3}\ \citeyear{#3},\ #2)}
\newcommand{\citec}[2]{\citeauthor{#1}\ (#2 \citeyear{#1})}
\DeclareUnicodeCharacter{FF08}{}

\begin{document}

\maketitle

\begin{abstract}
Networked data, in which every training example involves two objects and may share some common objects with others, is used in many machine learning tasks such as learning to rank and link prediction. 
A challenge of learning from networked examples is that target values are not known for some pairs of objects. 
In this case, neither the classical i.i.d.\ assumption nor techniques based on complete U-statistics can be used. 
Most existing theoretical results of this problem only deal with the classical empirical risk minimization (ERM) principle that always weights every example equally, 
but this strategy leads to unsatisfactory bounds. 
We consider general weighted ERM and show new universal risk bounds for this problem. 
These new bounds naturally define an optimization problem which leads to appropriate weights for networked examples. 
Though this optimization problem is not convex in general, we devise a new fully polynomial-time approximation scheme (FPTAS) to solve it. 
\end{abstract}

\section{Introduction}
\label{sec:introduction}
\textit{``No man is an island, entire of itself ...''}, the beginning of a well-known poem by the 17th century English poet John Donne, might be able to explain why social networking websites are so popular. 
These social media not only make communications convenient and enrich our lives but also bring us data, of an unimaginable amount, that is intrinsically networked. 
Social network data nowadays is widely used in research on social science, network dynamics, and as an inevitable fate, data mining and machine learning \cite{scott2017social}. 
Similar examples of networked data such as traffic networks \cite{min2011real}, chemical interaction networks \cite{szklarczyk2014string}, citation networks \cite{dawson2014current} also abound throughout the machine learning world. 


Admittedly, many efforts have been made to design practical algorithms for learning from networked data, e.g., \cite{liben2007link,macskassy2007classification,li2016robust,garcia2016combining}. 
However, not many theoretical guarantees of these methods have been established, which is the main concern of this paper. More specifically, this paper deals with risk bounds of \emph{classifiers trained with networked data (\problemabbr{})
} whose goal is to train a classifier with examples in a data graph $G$. 
Every vertex of $G$ is an object and described by a feature vector $X\in \xspace{}$ that is drawn independently and identically (i.i.d.) from an unknown distribution, while every edge corresponds to a training example whose input is a pair of feature vectors $\pair{X,X'}$ of the two ends of this edge and whose target value $Y$ is in $\{0,1\}$.

A widely used principle to select a proper model from a hypothesis set is \emph{Empirical Risk Minimization (ERM)}. 
\citet{papa2016graph} establish risk bounds for ERM on complete data graphs, and the bounds are independent of the distribution of the data. 
These bounds are of the order $O(\log(n)/n)$, where $n$ is the number of vertices in the complete graph. 
However, in practice it is very likely that one cannot collect examples for all pairs of vertices and then $G$ is usually incomplete, thus techniques based on complete $U$-processes in \cite{papa2016graph} cannot be applied and the risk bounds of the order $O(\log(n)/n)$ are no longer valid in this setting. 
By generalizing the moment inequality for $U$-processes to the case of incomplete graphs, we prove novel risk bounds for the incomplete graph.  

Usually, every training example is equally weighted (or unweighted) in ERM, which seems much less persuasive when the examples are networked, in particular when the graph is incomplete. But, most existing theoretical results of learning from networked examples are based on the unweighted ERM \cite{Usunier2005,ralaivola2009chromatic}, and their bounds are of the order $O(\sqrt{\chi^*(D_G)/m})$ where $D_G$ is the \emph{line graph} of $G$ and $\chi^*$ is the \emph{fractional chromatic number} of $D_G$ (see Section~A in the online appendix\footnote{https://arxiv.org/abs/1711.04297}) and $m$ is the number of training examples. 
In order to improve this bound, \citet{wang2017learning} propose \emph{weighted} ERM 
which adds weights to training examples 
according to the data graph, and show that the risk bound for weighted ERM can be of the order $O(1/\sqrt{\nu^*(G)})$ where $\nu^*(G)$ is the \emph{fractional matching number} of $G$, so using weighted ERM networked data can be more effectively exploited than the equal weighting method, as basic graph theory tells us $\nu^*_G\ge m / \chi^*(D_G)$. 
However, \citet{wang2017learning} (in fact, \citet{Usunier2005} and \citet{ralaivola2009chromatic} also) assume that any two examples can be arbitrarily correlated if they share a vertex, which cannot lead to an $O(\log(n)/n)$ bound when the graph is complete. 

We show that the \emph{``low-noise'' condition}, also called the \emph{Mammen-Tsybakov noise condition}  \cite{Mammen1998Smooth}, which is commonly assumed in many typical learning problems with networked data, e.g., ranking \cite{clemenccon2008ranking} and graph reconstruction \cite{papa2016graph}, can be used to reasonably bound the dependencies of two examples that share a common vertex and then leads to \emph{tighter} risk bounds. 

In summary, in this paper we mainly 
\begin{itemize}
\item prove new universal risk bounds for \problemabbr{} which
\begin{itemize}
\item can be applied to learning from networked data even if the data graph is incomplete; 
\item exploit the property of the ``low-noise'' condition, and then become tighter than previous results; 
\item allow non-identical weights on different examples, so it is possible to achieve better learning guarantee by choosing these weights.
\end{itemize}
\item formulate a non-convex optimization problem inspired by our new risk bounds (because our risk bounds depend on the weights added to every training example, and a better weighting scheme leads to a tighter bound), and then we also design a new efficient algorithm to obtain an approximate optimal weighting vector and show that this algorithm is a fully polynomial-time approximation scheme for this non-convex program. 
\end{itemize}

\begin{table*}[!htbp]
  \caption{\label{ta:methods}Summary of methods for \problemabbr{}.}
  \centering
  \begin{tabular}{M{2.7cm} M{2cm} M{4cm} M{3.7cm}}
    \toprule
    \textbf{Principles} & \textbf{Graph type} & \textbf{With ``low-noise'' condition} & \textbf{Without ``low-noise'' condition}\\
    \midrule
    \multirow{4}{*}{\parbox[t]{4cm}{\textbf{Unweighted ERM \\(equally weighted)}}} & 
    \textbf{Complete graphs} & \parbox[t]{4cm}{\citec{clemenccon2008ranking}{Ann.\ Stat.},\\ \citec{papa2016graph}{NIPS}} & \citec{Biau2006}{Statistics and Decisions}\\
    \cmidrule{2-4}
    & \textbf{General graphs} & \citec{DBLP:conf/icml/RalaivolaA15}{ICML} & \parbox[t]{4cm}{\citec{Usunier2005}{NIPS},\\ \citec{ralaivola2009chromatic}{AISTATS}}\\
    \cmidrule{1-4}
    \textbf{Weighted ERM} & \textbf{General graphs} & \cellcolor{Gray}\emph{This paper}  & \citec{wang2017learning}{ALT}\\
    \bottomrule
  \end{tabular}
\end{table*}

\section{Intuitions} 
\label{sub:intiutions}

We now have a look at previous works that are closely related to our work, as shown in Table~\ref{ta:methods}, and present the merits of our method. 
\citet{Biau2006}, \citet{clemenccon2008ranking} and \citet{papa2016graph} deal with the case when the graph is complete, i.e., the target value of every pair of vertices is known. 
In this case, \citet{clemenccon2008ranking} formulate the ``low-noise'' condition for the ranking problem and demonstrate that this condition can lead to tighter risk bounds by the moment inequality for $U$-processes.
\citet{papa2016graph} further consider the graph reconstruction problem introduced by \citet{Biau2006} and show this problem always satisfies the ``low-noise'' condition. 

If the graph is incomplete, one can use either Janson's decomposition \cite{janson2004large,Usunier2005,ralaivola2009chromatic,DBLP:conf/icml/RalaivolaA15} or the fractional matching approach by \citet{wang2017learning} to derive risk bounds. 
The main differences between these two approaches are: 
\begin{itemize}
  \item \citet{wang2017learning} consider the data graph $G$ while Janson's decomposition uses only the line graph $D_G$. 
  \item The fractional matching approach considers weighted ERM while \citet{janson2004large}, \citet{Usunier2005}, \citet{ralaivola2009chromatic} and \citet{DBLP:conf/icml/RalaivolaA15} only prove bounds for unweighted ERM.
\end{itemize}

Though \citet{wang2017learning} show improved risk bounds, as far as we know, there is no known tight risk bound on incomplete graphs for tasks such as pairwise ranking and graph reconstruction that satisfy the ``low-noise'' condition. Under this condition, the method proposed in \cite{wang2017learning} does not work (see Section~\ref{sub:complete_graph}). 



Before we show new risk bounds and new weighting methods, we present the following three aspects to convey some intuitions. 

\paragraph{Line Graphs} 
Compared to Janson's decomposition which is based on line graphs, our method utilizes the additional dependency information in the data graph $G$. 
For example, the complete line graph with three vertices (i.e., triangle) corresponds to two different data graphs, as illuminated in Figure~\ref{fig:weighting_scheme_complete_graph}. 
Hence, line graph based methods ignore some important information in the data graph. 
This negligence makes it unable to improve bounds, no matter whether considering weighted ERM or not (see Section~A.1 in the online appendix). 
In Section~\ref{sub:equally_weighting}, we show that our bounds are tighter than that of line graph based methods. 

\paragraph{Asymptotic Risk}
As mentioned by \citet{wang2017learning}, if several examples share a vertex, then we are likely to put less weight on them because the influence of this vertex to the empirical risk should be bounded. 
Otherwise, if we treat every example equally, then these dependent examples may dominate the training process and lead to the risk bounds that do not converge to $0$ (see the example in Section~\ref{sub:equally_weighting}).

\paragraph{Uniform Bounds} 
\citet{DBLP:conf/icml/RalaivolaA15} prove an entropy-base concentration inequality for networked data using Janson's decomposition, but the 
assumption there is usually too restrictive to be satisfied 
(see Section~A.2 in the online appendix). 
To circumvent this problem, our method uses 
the ``low-noise'' condition (also used in \cite{papa2016graph}) to establish uniform bounds, 
in absence of any restrictive condition imposed on the data distribution. 

\section{Preliminaries} 
\label{sec:preliminaries}

In this section, we begin with the detailed probabilistic framework for \problemabbr{}, and then give the definition of weighted ERM on networked examples. 

\subsection{Problem Statement} 
\label{sub:problem_statement}

Consider a graph $G=(V,E)$ with a vertex set $V=\{1,\dots,n\}$ and a set of edges $E\subseteq\{\edge{i,j}: 1\le i\neq j\le n\}$. 
For each $i\in V$, a continuous random variable (r.v.) $X_i$, taking its values in a measurable space $\xspace{}$, describes features of vertex $i$. The $X_i$'s are i.i.d.\ r.v.'s following some unknown distribution $P_\xspace$.
Each pair of vertices $\pair{i,j}\in E$ corresponds to a networked example whose \emph{input} is a pair $(X_i,X_j)$ and \emph{target} value is $Y_{i,j}\in\yspace{}.$ 
We focus on \emph{binary classification} in this paper, i.e., $\yspace{} = \{0,1\}$. 
Moreover, the distribution of target values only depends on the features of the vertices it contains but does not depend on features of other vertices, that is, there is a probability distribution $P_{\yspace{}\mid\xspace{}^2}$ such that for every pair $\pair{i,j}\in E$, the conditional probability 
\[P\left[Y_{i,j}=y\mid x_1,\ldots,x_n\right]=P_{\yspace{}\mid\xspace{}^2}\left[y,x_i,x_j\right].\]  

\begin{example}[pairwise ranking]
\cite{liu2009learning} categorize ranking problems into three groups by their input representations and loss functions. One of these categories is \emph{pairwise ranking} that learns a binary classifier telling which document is better in a given pair of documents. A document can be described by a feature vector from the $\xspace$ describing title, volume, \ldots The target value (rank) between two documents, that only depends on features of these two documents, is $1$ if the first document is considered better than the second, and $0$ otherwise. 
\end{example}

The training set $\trainingset:=\{(X_i,X_j,Y_{i,j})\}_{\pair{i,j}\in E}$ is \emph{dependent}\ copies of a generic random vector $(X_1, X_2, Y_{1,2})$ whose distribution $\distribution{}=P_\xspace\otimes P_\xspace\otimes P_{\yspace{}\mid\xspace{}^2}$ is fully determined by the pair $(P_\xspace{}, P_{\yspace{}\mid\xspace{}^2})$. 
Let $\mathcal{R}$ be the set of all measurable functions from $\xspace{}^2$ to $\yspace{}$ and for all $r\in\mathcal{R}$, the \emph{loss function} $\ell(r,(x_1,x_2,y_{1,2}))=\indicator_{y_{1,2}\neq r(x_1,x_2)}$. 

Given a graph $G$ with training examples $\trainingset$ and a \emph{hypothesis set}  $R\subseteq\mathcal{R}$, the \problemabbr{} problem is to find a function $r\in R$, with risk
\begin{equation}
    \label{eq:graph_reconstruction_risk}
    \risk(r) := \E[\ell(r, (X_1,X_2,Y_{1,2}))]
\end{equation}
that achieves a comparable performance to the \emph{Bayes rule} $r^*=\arg\inf_{r\in\mathcal{R}}\risk(r) = \indicator_{\eta(x_1,x_2)\ge 1/2}$, whose risk is denoted by $\bayeserror{}$, where $\eta(x_1,x_2)=P_{\yspace{}\mid\xspace{}^2}[1,x_1,x_2]$ is the \emph{regression function}. 

The main purpose of this paper is to devise a principle to select a classifier $\hat{r}$ from the hypothesis set $R$ and establish bounds for its \emph{excess risk} $\risk(\hat{r}) - \bayeserror{}$.
\begin{definition}[``low-noise'' condition]
  \label{de:low-noise condition}
  Let us consider a learning problem, in which the hypothesis set is $\mathcal{F}$ and the Bayes rule is $f^*$. With slightly abusing the notation, this problem satisfies the ``low-noise'' condition if $\forall f\in \mathcal{F}, \risk{}(f) - \bayeserror{}\ge C^\theta(\E[|f-f^*|])^\theta$ where $C$ is a positive constant.
\end{definition}
As mentioned, the ``low-noise'' condition can lead to tighter risk bounds.
For this problem, we show that the ``low-noise'' condition for the i.i.d.\ part of the Hoeffding decomposition \cite{hoeffding1948class} of its excess risk can be always obtained if the problem is \emph{symmetric} (see Lemma~\ref{le:variant_control}).
\begin{definition}[symmetry]
    \label{con:symmetric}
    A learning problem is symmetric if for every $x_i,x_j\in \xspace{}$, $y_{i,j}\in \yspace{}$ and $r\in R$, $\lossf(r, (x_i,x_j,y_{i,j})) = \lossf(r, (x_j,x_i,y_{j,i}))$.
\end{definition}
Many typical learning problems are symmetric.
For example, pairwise ranking problem with symmetric functions $r$ in the sense that $r(X_1,X_2)=1-r(X_2,X_1)$ satisfies the symmetric condition. 
\LongVersion
The link prediction which predicts the occurrence of an linkage between the vertices $i$ and $j$ that only depends on $X_i$ and $X_j$ with symmetric function $r$ in the sense that $r(X_1,X_2)=r(X_2,X_1)$ also satisfies Assumption~\ref{con:symmetric}.
\LongVersionEnd

\subsection{Weighted ERM} 

\label{sub:weighted_empirical_risk_minimization}

ERM aims to find the function from a hypothesis set that minimizes the empirical estimator of \eqref{eq:graph_reconstruction_risk} on the training examples $\trainingset{}=\{(X_i,X_j,Y_{i,j})\}_{\pair{i,j}\in E}$:
\begin{equation}
    \label{eq:graph_reconstruction_non_weighted_erm}
    \empiricalrisk{m}(r) := \frac{1}{m}\sum_{\pair{i,j}\in E} \lossf(r, (X_i,X_j, Y_{i,j})).
\end{equation}
where $m$ is the number of training examples. 
In this paper, we consider its weighted version, in which we put weights on the examples and select the minimizer $r_\weight{}$ of the weighted empirical risk
\begin{equation}
    \label{eq:graph_reconstruction_weighted_erm}
    \empiricalrisk{\weight}(r) := \frac{1}{\normo{\weight}} \sum_{\pair{i,,j}\in E} w_{i,j} \lossf(r, (X_i,X_j, Y_{i,j}))
\end{equation}
where $\weight$ is a \emph{fractional matching} of $G$ and $\normo{\weight{}} > 0$. 
\begin{definition}[fractional matching]
Given a graph $G=(V,E)$, a \emph{fractional matching} $\weight$ is a non-negative 
vector $(w_{i,j})_{\pair{i,j}\in E}$ that for every vertex $i\in V, \sum_{j:\pair{i,j}\in E}w_{i,j}\le 1$.
\end{definition}

\LongVersion
We can also interpret \eqref{eq:graph_reconstruction_weighted_erm} as the expected empirical risk of subsampling training set with replacement conditional on a training set.
Assuming we subsample $N$ examples from training set with the probability $p_{i,j}=w_{i,j}/\normo{\weight{}}$ for the example $(X_i,X_j,Y_{i,j})$, then we have the expectation of empirical risk of this subsampling method conditional on training set
\begin{equation}
    \label{eq:subsampling_erm}
    N\sum_{\{i,,j\}\in E} p_{i,j} \lossf(r, (X_i,X_j, Y_{i,j}))
\end{equation}
minimizing which is equivalent to minimizing \eqref{eq:graph_reconstruction_weighted_erm}.
\LongVersionEnd

\LongVersion
The empirical risk minimizer $r_\weight{}$ is a solution of the optimization problem $\inf_{r\in R}L_\weight{}(r)$ whose performance is measured by its excess risk $L(r_\weight{})-\bayeserror{}$.
If for all $\pair{i,j}\in E$, $w_{i,j}=1/|E|$, weighted ERM is degenerated to the classical ERM.
Let $E_<=\{\pair{i,j}: i<j, \pair{i,j}\in E\}$ be the ordered edge set. If Condition~\ref{con:symmetric} is satisfied, we also have
\begin{equation}
    \label{eq:symmetric_graph_reconstruction_imcomplete_erm}
    \empiricalrisk{\weight{}}(r) = \frac{2}{\normo{\weight{}}} \sum_{\pair{i,j}\in E_<} w_{i,j} \lossf(r, (X_i,X_j, Y_{i,j})).
\end{equation}
\LongVersionEnd

\section{Universal Risk Bounds} 
\label{sec:risk_bounds}
In this section, we use covering numbers as the complexity measurement of hypothesis sets to prove that tighter universal risk bounds are always attained by the minimizers of the weighted empirical risk \eqref{eq:graph_reconstruction_weighted_erm}.

\subsection{Covering Numbers} 
\label{sub:covering number}
The excess risk $\risk{}(r_\weight{})-\bayeserror{}$ depends on the hypothesis set $R$ whose complexity can be measured by \emph{covering number} \cite{cucker2007learning}. 
A similar but looser result using VC-dimension \cite{vapnik1971uniform} can be obtained as well. 

\begin{definition}[covering numbers]
Let $(\mathcal{F}, \lnorm{}_p)$ be a metric space with $\lnorm{}_p$-pseudometric. We define the covering number $N(\mathcal{F},\lnorm{}_p,\epsilon)$ be the minimal $l\in\mathbb{N}$ such that there exist $l$ disks in $\mathcal{F}$ with radius $\epsilon$ covering $\mathcal{F}$. If the context is clear, we simply denote $N(\mathcal{F},\lnorm{}_p, \epsilon)$ by $N_p(\mathcal{F},\epsilon)$.


\end{definition}

In this paper, we focus on the $\lnorm_\infty$ covering number $N_\infty(\mathcal{F},\epsilon)$ and suppose that it satisfies the following assumption.

\begin{assumption}\label{ass:covering_number} 
There exists a nonnegative number $\complexbound < 1$ and a constant $K$ such that $\log N_\infty(\mathcal{F},\epsilon) \le K\epsilon^{-\complexbound}$ for all $\epsilon\in (0,1]$. 
\end{assumption}

Similar to \cite{Massart2006} and \cite{rejchel2012ranking}, we restrict to $\complexbound{} < 1$, whereas in the empirical process theory this exponent usually belongs to $[0,2)$. This restriction is needed to prove Lemma~\ref{le:uniform_approximation}, which involves the integral of $\log N_\infty(\mathcal{F},\epsilon)$ through $0$. 
\citet{Dudley1974Metric}, \citet{Korostelev1993Minimax} and \citet{Mammen1995Asymptotical} presented various examples of classes $\mathcal{F}$ satisfying Assumption \ref{ass:covering_number}.
We also refer interested readers to (\citeauthor{Mammen1998Smooth} \citeyear{Mammen1998Smooth}, p.\ 1813) for more concrete examples of hypothesis classes with smooth boundaries satisfying Assumption~\ref{ass:covering_number}. 

\subsection{Risk Bounds} 
\label{sub:risk_bounds}
Now we are ready to show the tighter risk bounds for weighted empirical risk by the following theorem.
\begin{theorem}[risk bounds]
    \label{th:main_theorem}
    Let $r_\weight{}$ be a minimizer of the weighted empirical risk $\empiricalrisk{\weight{}}$ over a class $R$ that satisfies Assumption~\ref{ass:covering_number}. There exists a constant $C>0$ such that for all $\delta\in(0,1]$, with probability at least $1-\delta$, the excess risk of $r_\weight{}$ satisfies
    \DoubleColumn
    \begin{equation}
        \begin{aligned}
            \label{eq:main_result}
            \risk(r_\weight{}&)-\bayeserror \le 2(\inf_{r\in R}\risk(r)-\bayeserror) + \frac{K'C\log(1/\delta)}{(1-\complexbound)^{2/(\complexbound+1)}\normo{\weight{}}}\\
            &\bigg(\normo{\weight{}}^{\complexbound/(1+\complexbound)}+\max\Big(\|\weight{}\|_2,\|\weight{}\|_{\max}(\log(1/\delta))^{1/2},\\
            &\qquad\qquad\qquad\qquad\qquad\|\weight{}\|_\infty(\log(1/\delta))\Big)\bigg)
        \end{aligned}
    \end{equation}
    \DoubleColumnEnd
    \SingleColumn
    \begin{equation}
        \begin{aligned}
            \label{eq:main_result}
            \risk(r_\weight{})-\bayeserror &\le 2(\inf_{r\in R}\risk(r)-\bayeserror) + \frac{K'C\log(1/\delta)}{(1-\complexbound)^{2/(\complexbound+1)}\normo{\weight{}}}\bigg(\normo{\weight{}}^{\complexbound/(1+\complexbound)}\\
            &\qquad\qquad+\max\Big(\|\weight{}\|_2,\|\weight{}\|_{\max}(\log(1/\delta))^{1/2},\|\weight{}\|_\infty(\log(1/\delta))\Big)\bigg)
        \end{aligned}
    \end{equation}
    \SingleColumnEnd
    where $\|\weight{}\|_{\max}=\max_i \sqrt{\sum_{j:\pair{i,j}\in E} w_{i,j}^2}$ and $K'=\max(K,\sqrt{K},K^{1/(1+\complexbound{})})$.
\end{theorem}

According to Theorem~\ref{th:main_theorem}, if the parameter 
$\delta$ is greater than the value $\exp\left(-\min(\|\weight{}\|_2/\|\weight{}\|_\infty,\|\weight{}\|_2^2/\|\weight{}\|_\max^2)\right),$ 
then the risk bounds above are of the order
$O\left((1/\normo{\weight{}})^{1/(1+\complexbound)}+\|\weight{}\|_2/\normo{\weight{}}\right).$ 
In this case, our bounds are tighter than $O(1/\sqrt{\normo{\weight{}}})$ as $\|\weight\|_2/\normo{\weight}\le 1/\sqrt{\normo{\weight}}$ (recall that $\weight{}$ must be a fractional matching and $0<\complexbound{}<1$).
If $G$ is complete and every example is equally weighted, 
the bounds of the order $O((1/n)^{1/(1+\complexbound)})$ achieve the same results as in \cite{papa2016graph}\footnote{They consider the same range of $\delta$.} 

\begin{remark*}
Theorem~\ref{th:main_theorem} provides universal risk bounds no matter what the distribution of the data is. The factor of $2$ in front of the approximation error $\inf_{r\in R}\risk(r)-\bayeserror$ has no special meaning and can be replaced by any constant larger than $1$ with a cost of increasing the constant $C$. 
\citet{wang2017learning} obtain risk bounds that has a factor $1$ in front of the approximation error part, but in their result the bound is $O(1/\sqrt{\normo{\weight{}}})$. 
Hence, Theorem~\ref{th:main_theorem} improves their results if the approximation error does not dominate the other terms in the bounds.
\end{remark*}


In the rest of this section, we outline the main ideas to obtain this result.
We first define
\[q_r(x_1,x_2,y_{1,2}):=\lossf(r, x_1,x_2,y_{1,2})-\lossf(r^*,x_1,x_2,y_{1,2})\]
for every $(x_1,x_2,y_{1,2})\in \mathcal X\times\mathcal X\times\yspace{}$ and let $\Lambda(r) := \risk(r)-\bayeserror=\E[q_r(X_1,X_2,Y_{1,2})]$
be the excess risk with respect to the Bayes rule.
Its empirical estimate by weighted ERM is 
\DoubleColumn
\begin{align*}
    \Lambda_\weight{}(r) &= \empiricalrisk{\weight{}}(r)-\empiricalrisk{\weight{}}(r^*)\\
    &=\frac{1}{\normo{\weight{}}}\sum_{\pair{i,j}\in E} w_{i,j} q_r(X_i,X_j,Y_{i,j}).
\end{align*}
\DoubleColumnEnd
\SingleColumn
\begin{align*}
    \Lambda_\weight{}(r) = \empiricalrisk{\weight{}}(r)-\empiricalrisk{\weight{}}(r^*)=\frac{1}{\normo{\weight{}}}\sum_{\pair{i,j}\in E} w_{i,j} q_r(X_i,X_j,Y_{i,j}).
\end{align*}
\SingleColumnEnd

By Hoeffding's decomposition \cite{hoeffding1948class}, for all $r\in\mathcal R$, one can write
\begin{equation}
    \label{eq:graph_reconstruction_hoeffding_decomposition}
    \Lambda_\weight{}(r) = T_\weight{}(r) + U_\weight{}(r) + \widetilde{U}_\weight{}(r),
\end{equation}
where
\[T_\weight{}(r)=\Lambda(r)+\frac{2}{\normo{\weight{}}}\sum_{i=1}^n\sum_{j:\pair{i,j}\in E}w_{i,j}h_r(X_i)\]
is a weighted average of i.i.d.\ random variables with $h_r(X_i)=\E[q_r(X_i,X_j,Y_{i,j})\mid X_i]-\Lambda(r)$,
\[U_\weight{}(r)=\frac{1}{\normo{\weight{}}}\sum_{\pair{i,j}\in E} w_{i,j}(\hat{h}_r(X_i,X_j)\]
is a weighted \emph{degenerated} (i.e., the symmetric kernel $\hat{h}_r(x_1,x_2)$ such that $\E[\hat{h}_r(X_1, X_2)\mid X_1=x_1]=0$ for all $x_1\in\xspace{}$) $U$-statistic 
$\hat{h}_r(X_i,X_j)=\E[q_r(X_i,X_j,Y_{i,j})\mid X_i,X_j]-\Lambda(r)-h_r(X_i)-h_r(X_j)$ 
and
\[\widetilde{U}_\weight{}(r)=\frac{1}{\normo{\weight{}}}\sum_{\pair{i,j}\in E}w_{i,j}\tilde{h}_r(X_i,X_j,Y_{i,j})\]
with a \emph{degenerated} kernel 
$\tilde{h}_r(X_i,X_j,Y_{i,j})=q_r(X_i,X_j\linebreak,Y_{i,j})-\E[q_r(X_i,X_j,Y_{i,j})\mid X_i,X_j].$ 
In the following, we bound the three terms $T_\weight{}, U_\weight{}$ and $\widetilde{U}_\weight{}$ in \eqref{eq:graph_reconstruction_hoeffding_decomposition} respectively. 


\begin{lemma}[uniform approximation]
    \label{le:uniform_approximation}
    Under the same assumptions as in Theorem~\ref{th:main_theorem}, for any $\delta\in (0,1/e)$, we have with probability at least $1-\delta$,
    \SingleColumn
    \[\sup_{r\in R}|U_\weight{}(r)|\le \frac{\max(K,\sqrt{K})C_1}{1-\complexbound{}}\max\Big(\frac{\|\weight{}\|_2\log(1/\delta)}{\normo{\weight{}}},\frac{\|\weight{}\|_{\max}(\log(1/\delta))^{3/2}}{\normo{\weight{}}}, \frac{\|\weight{}\|_\infty(\log(1/\delta))^2}{\normo{\weight{}}}\Big)\]
    \SingleColumnEnd
    \DoubleColumn
    \begin{align*}
        \sup_{r\in R}|U_\weight{}(r)|\le \frac{\max(K,\sqrt{K})C_1}{1-\complexbound{}}\max\Big(\frac{\|\weight{}\|_2\log(1/\delta)}{\normo{\weight{}}},\\
        \frac{\|\weight{}\|_{\max}(\log(1/\delta))^{3/2}}{\normo{\weight{}}}, \frac{\|\weight{}\|_\infty(\log(1/\delta))^2}{\normo{\weight{}}}\Big)
    \end{align*}
    \DoubleColumnEnd
    and
    \SingleColumn
    \[\sup_{r\in R}|\widetilde{U}_\weight{}(r)|\le \frac{\max(K,\sqrt{K})C_2}{1-\complexbound{}}\bigg(\frac{\|\weight{}\|_2}{\normo{\weight{}}}+\max\Big(\frac{\|\weight{}\|_{\max}(\log(1/\delta))^{3/2}}{\normo{\weight{}}},\frac{\|\weight{}\|_\infty(\log(1/\delta))^2}{\normo{\weight{}}}\Big)\bigg)\]
    \SingleColumnEnd
    \DoubleColumn
    \begin{align*}
        \sup_{r\in R}&|\widetilde{U}_\weight{}(r)|\le \frac{\max(K,\sqrt{K})C_2}{1-\complexbound{}}\bigg(\frac{\|\weight{}\|_2}{\normo{\weight{}}}\\
        &+\max\Big(\frac{\|\weight{}\|_{\max}(\log(1/\delta))^{3/2}}{\normo{\weight{}}},\frac{\|\weight{}\|_\infty(\log(1/\delta))^2}{\normo{\weight{}}}\Big)\bigg)
    \end{align*}
    \DoubleColumnEnd
    where $C_1,C_2 < +\infty$ are constants.
\end{lemma}

To prove Lemma~\ref{le:uniform_approximation}, we show that $U_\weight{}(r)$ and $\widetilde{U}_\weight{}(r)$ can be bounded by Rademacher chaos using classical symmetrization and randomization tricks combined with the decoupling method. We handle these Rademacher chaos by generalizing the moment inequality for $U$-statistics in \cite{clemenccon2008ranking}. 
Specifically, we utilize the moment inequalities from \cite{Boucheron2005} to convert them into a sum of simpler processes, which can be bounded by the metric entropy inequality for Khinchine-type processes \citep{see}{Proposition 2.6}{Dembo1994} and Assumption~\ref{ass:covering_number}. The detailed proofs can be found in Section~C in the online appendix.

Lemma~\ref{le:uniform_approximation} shows that the contribution of the degenerated parts $U_\weight{}(r)$ and $\widetilde{U}_\weight{}(r)$ to the excess risk can be bounded. 
This implies that minimizing $\Lambda_\weight{}(r)$ is approximately equivalent to minimizing $T_\weight{}(r)$ and thus $r_\weight{}$ is a $\rho$-minimizer of $T_\weight{}(r)$ in the sense that $T_\weight{}(r_\weight{})\le \rho+\inf_{r\in R}T_\weight{}(r)$.
In order to analyze $T_\weight{}(r)$, which can be treated as a weighted empirical risk on i.i.d.\ examples, we generalize the results in \cite{Massart2006} (see Section~B in the online appendix). 
Based on this result, tight bounds for the excess risk with respect to $T_\weight{}(r)$
can be obtained if the variance of the excess risk is controlled by its expected value.
By Lemma~\ref{le:variant_control}, $T_\weight{}(r)$ fulfills this condition, which leads to Lemma~\ref{le:risk_bounds_iid}.

\begin{lemma}[condition leads to ``low-noise'', {\cite[Lemma 2]{papa2016graph}}]
    \label{le:variant_control}
    If the learning problem \problemabbr{} is symmetric, then
    \begin{equation}
        \label{eq:variant_control}
        \Var\left[\E[q_r(X_1,X_2,Y_{1,2})\mid X_1]\right]\le \Lambda(r) 
    \end{equation}
    holds for any distribution $\distribution$ and any function $r\in R$.
\end{lemma}

\begin{lemma}[risk bounds for i.i.d.\ examples]
\label{le:risk_bounds_iid}
  Suppose that $r'$ is a $\rho$-minimizer of $T_\weight{}(r)$ in the sense that $T_\weight{}(r')\le \rho + \inf_{r\in R}T_\weight{}(r)$ and $R$ satisfies Assumption~\ref{ass:covering_number}, then there exists a constant $C$ such that for all $\delta\in(0,1]$, with probability at least $1-\delta$, the risk of $r'$ satisfies
  \[\Lambda(r')\le 2\inf_{r\in R}\Lambda(r)+2\rho+\frac{CK^{1/(1+\complexbound{})}\log(1/\delta)}{(\normo{\weight{}}(1-\complexbound{})^2)^{1/(1+\complexbound{})}}.\]
\end{lemma}

With Lemma \ref{le:uniform_approximation} Lemma \ref{le:risk_bounds_iid}, now we are ready to prove Theorem \ref{th:main_theorem}.

\begin{proof}[Proof of Theorem~\ref{th:main_theorem}]
    Let us consider the Hoeffding decomposition \eqref{eq:graph_reconstruction_hoeffding_decomposition} of $\Lambda_\weight{}(r)$ that is minimized over $r\in R$. The idea of this proof is that the degenerate parts $U_\weight{}(r)$ and $\widetilde{U}_\weight{}(r)$ can be bounded by Lemma~\ref{le:uniform_approximation}. Therefore, $r_\weight{}$ is an approximate minimizer of $T_\weight{}(r)$, which can be handled by Lemma~\ref{le:risk_bounds_iid}.

    Let $A$ be the event that
    \[\sup_{r\in R}|U_\weight{}(r)|\le \kappa_1,\]
    where
    \DoubleColumn
    \begin{align*}
    \kappa_1&=\frac{C_1}{1-\complexbound{}}\max\Big(\frac{\|\weight{}\|_2\log(1/\delta)}{\normo{\weight{}}},\frac{\|\weight{}\|_{\max}(\log(1/\delta))^{3/2}}{\normo{\weight{}}},\\
    &\qquad\qquad \frac{\|\weight{}\|_\infty(\log(1/\delta))^2}{\normo{\weight{}}}\Big)
    \end{align*}
    \DoubleColumnEnd
    \SingleColumn
    \begin{align*}
    \kappa_1&=\frac{C_1}{1-\complexbound{}}\max\Big(\frac{\|\weight{}\|_2\log(1/\delta)}{\normo{\weight{}}},\frac{\|\weight{}\|_{\max}(\log(1/\delta))^{3/2}}{\normo{\weight{}}}, \frac{\|\weight{}\|_\infty(\log(1/\delta))^2}{\normo{\weight{}}}\Big)
    \end{align*}
    \SingleColumnEnd
    for an appropriate constant $C_1$. Then by Lemma~\ref{le:uniform_approximation}, $\Pro[A]\ge 1-\delta/4$. Similarly, let $B$ be the event that
    \[\sup_{r\in R}|\widetilde{U}_\weight{}(r)|\le \kappa_2.\]
    where
    \DoubleColumn
    \begin{align*}
    \kappa_2&=\frac{C_2}{1-\complexbound{}}\bigg(\frac{\|\weight{}\|_2}{\normo{\weight{}}}+\max\Big(\frac{\|\weight{}\|_{\max}(\log(1/\delta))^{3/2}}{\normo{\weight{}}},\\
    &\qquad\qquad\frac{\|\weight{}\|_\infty(\log(1/\delta))^2}{\normo{\weight{}}}\Big)\bigg)
    \end{align*}
    \DoubleColumnEnd
    \SingleColumn
    \begin{align*}
    \kappa_2&=\frac{C_2}{1-\complexbound{}}\bigg(\frac{\|\weight{}\|_2}{\normo{\weight{}}}+\max\Big(\frac{\|\weight{}\|_{\max}(\log(1/\delta))^{3/2}}{\normo{\weight{}}},\frac{\|\weight{}\|_\infty(\log(1/\delta))^2}{\normo{\weight{}}}\Big)\bigg)
    \end{align*}
    \SingleColumnEnd
    for an appropriate constant $C_2$.
    Then $\Pro[B]\ge 1-\delta/4$.

    By \eqref{eq:graph_reconstruction_hoeffding_decomposition}, it is clear that, if both $A$ and $B$ happen,
    $r_\weight{}$ is a $\rho$-minimizer of $T_\weight{}(r)$ over $r\in R$ in the sense that the difference between the value of this latter quantity at its minimum and $r_\weight{}$ is at most $(\kappa_1+\kappa_2)$.
    Then, from Lemma~\ref{le:risk_bounds_iid}, with probability at least $1-\delta/2$, $r_\weight{}$ is a $(\kappa_1+\kappa_2)$-minimizer of $T_\weight{}(r)$, which the result follows. 
\end{proof}

An intuition obtained from our result is how to choose weights for networked data. By Theorem~\ref{th:main_theorem}, to obtain tight risk bounds, we need to maximize $\normo{\weight{}}$ (under the constraint that this weight vector is a fractional matching), which resembles the result of \cite{wang2017learning} (but they only need to maximize $\normo{\weight{}}$ and this is why they end in the $O(1/\sqrt{\nu^*(G)})$ bound), while making $\|\weight{}\|_2,\|\weight{}\|_\max,\|\weight{}\|_\infty$ as small as possible, which appears to suggest putting nearly average weights on examples and vertices respectively. These two objectives, maximizing $\normo{\weight{}}$ and minimizing $\|\weight{}\|_2,\|\weight{}\|_\max,\|\weight{}\|_\infty$, seem to contradict each other. 
In the next section, we discuss how to solve this problem.  


\LongVersion
Another advantage of our result is that if the risk bounds are optimized, we could reduce the computation complexity of minimizing $\empiricalrisk{\weight{}}$ by subsampling examples with large weights which by Theorem~\ref{th:main_theorem} may not significantly deteriorate the performance of weighted ERM.
\LongVersionEnd

\section{Weighting Vector Optimization} 
\label{sec:weighting_scheme}

In this section, we first formulate the optimization problem that minimizes the risk bounds in Theorem \ref{th:main_theorem}. 
Although this optimization problem is not convex unless $\beta=0$, which usually means that there is no general efficient way to solve it, we devise a \emph{fully polynomial-time approximation scheme (FPTAS)} to solve it.
\begin{definition}[FPTAS]
  An algorithm $\mathcal{A}$ is a FPTAS for a minimization problem $\Pi$, if for any input $\mathcal{I}$ of $\Pi$ and $\epsilon>0$, $\mathcal{A}$ finds a solution $s$ in time polynomial in both the size of $\mathcal{I}$ and $1/\epsilon$ that satisfies $f_\Pi(s)\le (1+\epsilon) \cdot f_\Pi(s^*)$,
  where $f_\Pi$ is the (positive) objective function of $\Pi$ and $s^*$ is an optimal solution for $\mathcal{I}$.
\end{definition}

\subsection{Optimization Problem} 
\label{sub:optimization_problem}
According to Theorem~\ref{th:main_theorem}, given a graph $G$, $\complexbound\in (0,1)$ and $\delta\in(0,1]$, one can find a good weighting vector with tight risk bounds by solving the following program:
\DoubleColumn
\begin{equation}
    \label{eq:optimization_program}
\begin{aligned}
    \min_{\weight{}} \quad&\frac{1}{\normo{\weight{}}}\bigg(\normo{\weight{}}^{\complexbound/(1+\complexbound)}+\max\Big(\|\weight{}\|_2,\\
    &\qquad\|\weight{}\|_{\max}(\log(1/\delta))^{1/2},
    \|\weight{}\|_\infty(\log(1/\delta))\Big)\bigg)\\
    \mbox{s.t.} \quad& \forall \pair{i,j}\in E, w_{i,j}\ge 0 \quad \mbox{and} \quad  \forall i, \sum_{j:\pair{i,j}\in E} w_{i,j}\le 1
\end{aligned}
\end{equation}
\DoubleColumnEnd
\SingleColumn
\begin{equation}
    \label{eq:optimization_program}
\begin{aligned}
    \min_{\weight{}} & \quad\frac{1}{\normo{\weight{}}}\bigg(\normo{\weight{}}^{\complexbound{}/(1+\complexbound{})}+\max\Big(\|\weight{}\|_2,\|\weight{}\|_{\max}(\log(1/\delta))^{1/2},\|\weight{}\|_\infty(\log(1/\delta))\Big)\bigg)\\
    \mbox{s.t.} \quad& \forall \pair{i,j}\in E, w_{i,j}\ge 0 \quad \mbox{and} \quad  \forall i, \sum_{j:\pair{i,j}\in E} w_{i,j}\le 1
\end{aligned}
\end{equation}
\SingleColumnEnd

To get rid of the fraction of norms in the program above, we consider a distribution $\probdistri{}$ on edges $p_{i,j}:=w_{i,j}/\normo{\weight{}}$ and then 
$\normo{\weight{}} \le 1/\max_{i=1,\dots,n} \sum_{j:\pair{i,j}\in E} p_{i,j}.$
Every distribution $\probdistri{}$ corresponds to a valid weighting vector $\weight{}$. 
By introducing two auxiliary variables $a$ and $b$, solving the original program \eqref{eq:optimization_program} is equivalent to solving
\SingleColumn
\begin{equation}
    \label{eq:optimization_program_1}
\begin{aligned}
    \min_{\probdistri{}} &\quad
    (\max_{i=1,\dots,n} \sum_{j:\pair{i,j}\in E} p_{i,j})^{1/(1+\complexbound{})}+\max\Big(\|\probdistri{}\|_2,\|\probdistri{}\|_{\max}(\log(1/\delta))^{1/2},
    \|\probdistri{}\|_\infty(\log(1/\delta))\Big)\\
    \mbox{s.t.} &\quad \forall \pair{i,j}\in E, p_{i,j}\ge 0 \quad \mbox{and} \quad \sum_{\pair{i,j}\in E} p_{i,j}= 1
\end{aligned}
\end{equation}
\SingleColumnEnd



\SingleColumn
\begin{equation}
    \label{eq:optimization_program_final}
\begin{aligned}
    \min_{a,b,\probdistri{}} \quad&
    a^{1/(1+\complexbound{})}+b\\
    \mbox{s.t.} \quad &\forall \pair{i,j}\in E, p_{i,j}\ge 0; \quad \forall \pair{i,j}\in E, p_{i,j}\log(1/\delta) - b \le 0\\
    & \forall i, \sum_{j:\pair{i,j}\in E} p_{i,j} -a  \le 0; \quad \forall i, \left(\sum_{j:\pair{i,j}\in E} p^2_{i,j}\log(1/\delta)\right)^{1/2} -b  \le 0\\
    & \|\probdistri{}\|_2 -b \le 0 \quad \mbox{and} \quad \sum_{\pair{i,j}\in E} p_{i,j}= 1
\end{aligned}
\end{equation}
\SingleColumnEnd
\DoubleColumn
\begin{equation}
    \label{eq:optimization_program_final}
\begin{aligned}
    \min_{a,b,\probdistri{}} \quad&
    a^{1/(1+\complexbound{})}+b\\
    \mbox{s.t.} \quad &\forall \pair{i,j}\in E, p_{i,j}\ge 0 \\
    & \forall \pair{i,j}\in E, p_{i,j}\log(1/\delta) - b \le 0\\
    & \forall i, \sum_{j:\pair{i,j}\in E} p_{i,j} -a  \le 0\\
    & \forall i, \left(\sum_{j:\pair{i,j}\in E} p^2_{i,j}\log(1/\delta)\right)^{1/2} -b  \le 0\\
    & \|\probdistri{}\|_2 -b \le 0 \quad \mbox{and} \quad \sum_{\pair{i,j}\in E} p_{i,j}= 1
\end{aligned}
\end{equation}
\DoubleColumnEnd
Note that the constraints are all convex. 
If $\complexbound{}=0$, e.g., the hypothesis set is finite, then the objective function becomes linear and thus \eqref{eq:optimization_program_final} is a convex optimization problem that can be solved by some convex optimization method (see e.g., \cite{boyd2004convex}) such as interior-point method. 

If $\complexbound{}>0$, the objective function is not convex any more. In fact, the program \eqref{eq:optimization_program_final} becomes a concave problem that may be optimized globally by some complex algorithms \cite{benson1995concave,hoffman1981method} that often need tremendous computation. 
Instead, one may only need to approximate it using some efficient methods, e.g., Concave-Convex Procedure \cite{Yuille2001The} and Coordinate Descent \cite{wright2015coordinate}. 
However, these methods lack in complexity analysis and may lead to a local optimum. 

\subsection{A Fully Polynomial-time Approximation Scheme} 
\label{sub:ptas}

To solve the program \eqref{eq:optimization_program_final} efficiently, we propose Algorithm \ref{alg:FPTAS} and show that it is a fully polynomial-time approximation scheme for \eqref{eq:optimization_program_final}. 

\begin{algorithm}[!htb] 
\caption{FPTAS for weighting vector optimization.} 
\label{alg:FPTAS} 
\begin{algorithmic}[1]
\Require 
$\epsilon$, $\beta$, $\delta$ and a graph $G$ that contains $n$ vertices and $m$ edges. 
\Ensure 
An approximate optimal weighting vector $\bar{\probdistri{}}$ for the program \eqref{eq:optimization_program_final}. 
\State Solve the following linear program (LP) efficiently\footnotemark,
and obtain an $\epsilon$-approximation 
$a_{min}$;
  \begin{equation}
    \label{eq:optimization_program_final_fix_b}
\begin{aligned}
    \min_{a,\probdistri{}} \quad&
    a\\
    \mbox{s.t.} \quad &\forall \pair{i,j}\in E, p_{i,j}\ge 0 \\
    & \forall i, \sum_{j:\pair{i,j}\in E} p_{i,j} -a  \le 0\\
    & \sum_{\pair{i,j}\in E} p_{i,j}= 1
\end{aligned}
\end{equation}
\State Let $Grid:=\{a_{min}+i\cdot\epsilon(1+\beta)/n \mid  i\in \mathbb{N} \hbox{ and } i\le n(1-a_{min})/\epsilon(1+\beta)\}$ and $Solutions := \emptyset$. 
\For{$a \in Grid$}
  \State Use some efficient interior-point method to obtain an $\epsilon$-approximation of the following program and add the solution $(a,b,\probdistri{})$ into $Solutions$.
  \begin{equation}
    \label{eq:optimization_program_final_fix_a}
\begin{aligned}
    \min_{b,\probdistri{}} \quad&
    b\\
    \mbox{s.t.} \quad &\forall \pair{i,j}\in E, p_{i,j}\ge 0 \\
    & \forall \pair{i,j}\in E, p_{i,j}\log(1/\delta) - b \le 0\\
    & \forall i, \sum_{j:\pair{i,j}\in E} p_{i,j} -a  \le 0\\
    & \forall i, \left(\sum_{j:\pair{i,j}\in E} p^2_{i,j}\log(1/\delta)\right)^{1/2} -b  \le 0\\
    & \|\probdistri{}\|_2 -b \le 0 \quad \mbox{and} \quad \sum_{\pair{i,j}\in E} p_{i,j}= 1
\end{aligned}
\end{equation}

\EndFor\\
\Return the vector $\bar{\probdistri{}}$ which makes $a^{1/(1+\beta)}+b$ smallest from $Solutions$. 
\end{algorithmic}
\end{algorithm}
\footnotetext{For example, some interior-point method.}

\begin{theorem}
\label{th:fptas}
  Algorithm \ref{alg:FPTAS} is a FPTAS for the program \eqref{eq:optimization_program_final}.
\end{theorem}

\begin{proof}
We first analyze the running time of this algorithm.   

  Note that $1/n\le a\le 1$ if the graph is not empty. 
  In Algorithm \ref{alg:FPTAS}, we first divide the problem into at most
  \[\frac{1-1/n}{\epsilon(1+\beta)/n}=\frac{n-1}{\epsilon(1+\complexbound{})}\]
  convex programs, each of which produces an $\epsilon$-approximate solution by some interior-point method. 
  Since interior-point method is FPTAS for convex problems \cite{boyd2004convex}, solving each of these programs needs polynomial time in the problem size $m+n$ and $1/\epsilon$. 
  Thus, the complexity of Algorithm \ref{alg:FPTAS} is also polynomial in $m+n$ and $1/\epsilon$.

Now we show that this algorithm indeed results in an $\epsilon$-approximation of this optimal solution. 

For any optimal solution $(a^*, b^*, \probdistri{}^*)$, 
if $a^*$ achieves minimum for the program 
\eqref{eq:optimization_program_final_fix_b}, we can find $a'$ in $Grid$ (actually $a_{min}$) such that 
  \begin{equation}
  \begin{aligned}
    \label{eq:a_mini_bound}
    (a')^{1/(1+\complexbound{})}&\le (1+\epsilon)^{1/(1+\complexbound{})}(a^*)^{1/(1+\complexbound{})}\\
    &\le (1+\epsilon)(a^*)^{1/(1+\complexbound{})}.
    \end{aligned}
  \end{equation}
  Otherwise, we can also find $a'$ in $Grid$ such that $a^*\le a'<a^*+\epsilon(1+\beta)/n$ and thus

  \begin{equation}
  \label{eq:a_bound}
  \begin{aligned}
    (a')^{1/(1+\complexbound{})}&\le (a^*+\epsilon(1+\beta)/n)^{1/(1+\complexbound{})}\\
    &\le (a^*)^{1/(1+\complexbound{})} + \epsilon(1+\beta)/n\\
    &\qquad\frac{1}{1+\complexbound{}}(a^*)^{-\complexbound{}/(1+\complexbound{})}\\
    &\le (1+\epsilon)(a^*)^{1/(1+\complexbound{})}
  \end{aligned}
  \end{equation}
  The third inequality follows from the fact that $1/n\le a^*$.
  We assume that the optimal solution for the program \eqref{eq:optimization_program_final_fix_a} is $b=b'$ when we fix $a=a'$.
  Because $(a^*,b^*)$ is feasible and $a'>a^*$, $(a', b^*)$ is always a feasible solution for the program \eqref{eq:optimization_program_final_fix_a}, which leads to $b'\le b^*$.
  Besides, interior-point method can produce an $\epsilon$-approximate solution $b''$ such that
  \begin{equation}
    \label{eq:b_bound}
    b''\le (1+\epsilon)b'\le (1+\epsilon)b^*.
  \end{equation}
  Finally, we select the best approximate weighting vector $\bar{\probdistri{}}$ from all solutions in $Solutions$. 
  Combining \eqref{eq:a_mini_bound}, \eqref{eq:a_bound} and \eqref{eq:b_bound}, we have the objective value for $\bar{\probdistri{}}$
  \begin{align*}
    (a_{\bar{\probdistri{}}})^{1/(1+\complexbound{})} + b_{\bar{\probdistri{}}} & \le (a')^{1/(1+\complexbound{})} + b''\\
    &\le (1+\epsilon)((a^*)^{1/(1+\complexbound{})} + b^*).
  \end{align*}
\end{proof}





\section{Discussion} 
\label{sec:discussion}
In this section, we first show that, according to our bounds, equal weighting is indeed the best weighting scheme for complete graphs. 
Then, we discuss the performance of this equal weighting scheme when the graph is incomplete. 

\subsection{Complete Graphs} 
\label{sub:complete_graph}
\begin{figure}[ht]
  \centering
  \subcapraggedrighttrue
  \DoubleColumn
    \subfigure[]{\label{fg:clique_1}
        \includegraphics[width=0.095\textwidth]{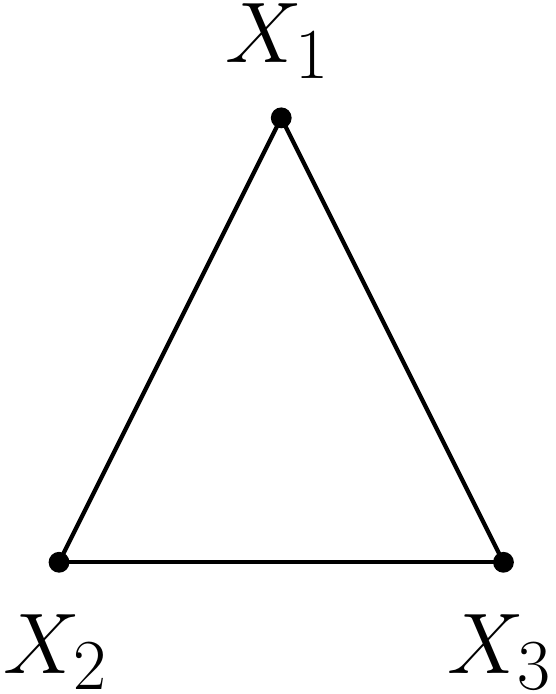}
    }\hspace{1ex}
    \subfigure[]{\label{fg:clique_2}
        \includegraphics[width=0.095\textwidth]{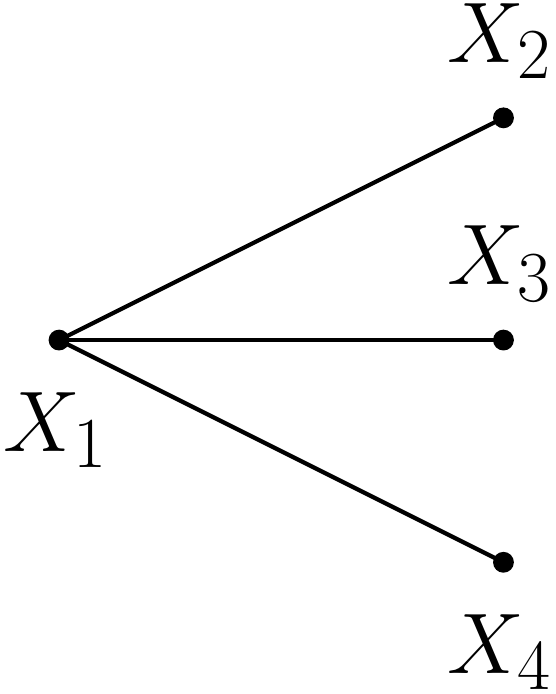}
    }\hspace{1ex}
    \subfigure[]{\label{fg:averaged_weighting}
        \includegraphics[width=0.095\textwidth]{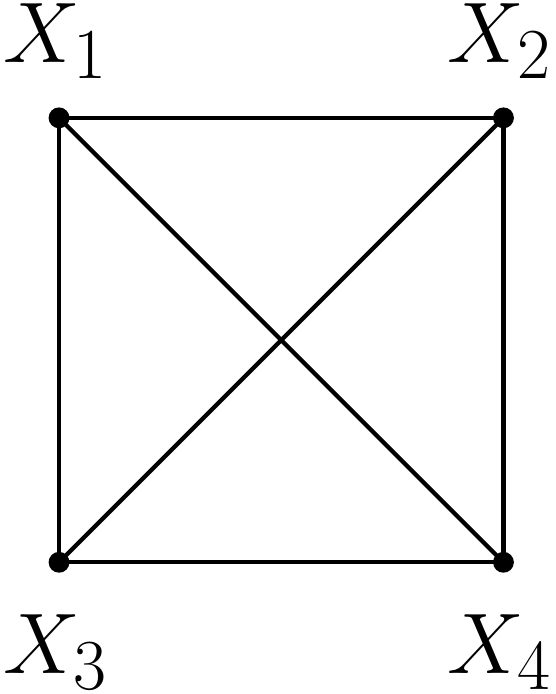}
    }\hspace{1ex}
    \subfigure[]{\label{fg:independent weighting}
        \includegraphics[width=0.095\textwidth]{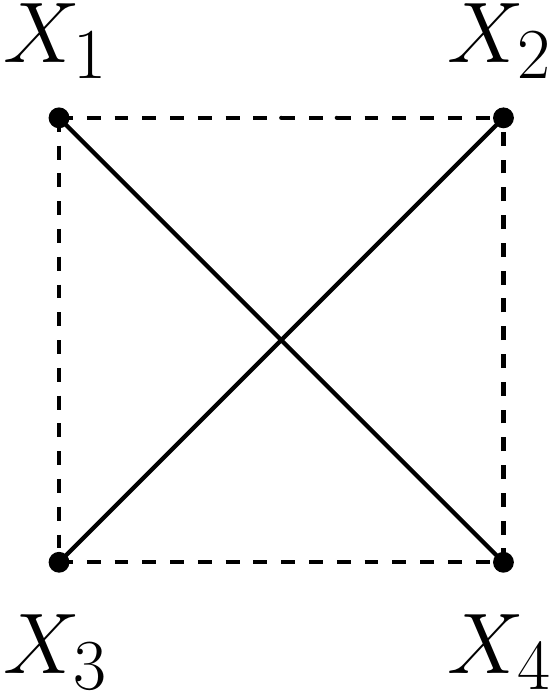}
    }
  \DoubleColumnEnd
  \SingleColumn
    \subfigure[First clique form]{\label{fg:clique_1}
        \includegraphics[width=0.17\textwidth]{clique_1.pdf}
    }\hspace{5ex}
    \subfigure[Second clique form]{\label{fg:clique_2}
        \includegraphics[width=0.17\textwidth]{clique_2.pdf}
    }\hspace{11ex}
    \subfigure[Equal weighting]{\label{fg:averaged_weighting}
        \includegraphics[width=0.17\textwidth]{complete_graph_1.pdf}
    }\hspace{5ex}
    \subfigure[Independent subset]{\label{fg:independent weighting}
        \includegraphics[width=0.17\textwidth]{complete_graph_2.pdf}
    }
  \SingleColumnEnd
    \caption{(a) and (b) are two different data graphs, but both of them correspond to the same line graph (a triangle). 
    (c) and (d) are two weighting schemes for a complete graph formed by points $X_1$, $X_2$, $X_3$, $X_4$. Solid line means its weight $p>0$ while dash line means $p=0$. (c): Weight every example equally. (d): Only the two examples in an independent subset get equally non-zero weights and other weights are $0$ (dashed line). Note that $\max_{i=1,\dots,n} \sum_{j:\pair{i,j}\in E} p_{i,j}$ of these two weighting schemes are the same, but (c) has the tighter risk bounds, as $\|\probdistri{}\|_2, \|\probdistri{}\|_\max$ and $\|\probdistri{}\|_\infty$ of (c) are smaller than that of (d) respectively.}
    \label{fig:weighting_scheme_complete_graph}
\end{figure}

When graph $G$ is complete, weighting all examples equally gives the best risk bound, as all the terms $\max_{i=1,\dots,n} \sum_{j:\pair{i,j}\in E} p_{i,j}$, $\|\probdistri{}\|_2$, $\|\probdistri{}\|_\max$ and $\|\probdistri{}\|_\infty$ achieve minimum. 
Compared to the results in \cite{wang2017learning}, our theory puts additional constrains on $\|\probdistri{}\|_2$, $\|\probdistri{}\|_\max$ and $\|\probdistri{}\|_\infty$ which encourages weighting examples fairly in this case, as illustrated in Figure~\ref{fig:weighting_scheme_complete_graph}. 
Besides, this scheme, which coincides with $U$-statistics that \emph{average} the basic estimator applied to all sub-samples, produces the smallest variance among all unbiased estimators \cite{hoeffding1948class}. 

\subsection{Equal Weighting} 
\label{sub:equally_weighting}
Let us discuss further the equal weighting scheme that gives every example the same weight. Denote by $\Delta(G)$ the maximum degree of $G$ (note that this is not the maximum degree of $D_G$) and let $p_{i,j}=1/m$ (recall that $m$ is the number of examples) for all $\pair{i,j}\in E$. According to program \eqref{eq:optimization_program_final},
using equal weighting scheme, the risk bounds are of the order 
\begin{equation}
\label{eq:equally_weighting_scheme}
    O\left((\frac{\Delta(G)}{m})^{1/(1+\complexbound{})} + \frac{1}{\sqrt{m}}\right),
\end{equation}
if $\delta\in (\exp(-m/\Delta(G)),1]$. In some cases, such as bounded degree graphs and complete graphs, this scheme provides reasonable risk bounds.
Note that $\Delta(G)$ is smaller than the maximum size of cliques in its corresponding line graph $D_G$ and $\fcoloring{}(D_G)$ is larger than the maximum size of cliques in $D_G$, these bounds above are always better than the bounds of the order $O(\sqrt{\fcoloring{}(D_G)/m})$ built by Janson's decomposition. 

However, as argued in Section~\ref{sub:intiutions},
one can construct examples to illustrate that if we use the equal weighting strategy when $\Delta(G)$ is large (e.g., if it is linear to $m$), the risk bounds \eqref{eq:equally_weighting_scheme} are very large and do not converge to $0$, while this problem can be solved by simply using a better weighting strategy. 
\begin{example}
Consider a data graph with $|E|=m \gg 1$ and $E$ consists of $m/2$ disjoint edges and $m/2$ edges sharing a common vertex, then $\Delta(G)=m/2$. Using the equal weighting scheme, the risk bounds are of the order $O(1)$ that is meaningless. A much better weighting scheme of this case is to weight the examples of disjoint edges with $2/(m+2)$ while weight the examples of adjacent edges with $4/m(m+2)$, which provides risk bounds of the order $O\left((1/m)^{1/(1+\complexbound{})} + \sqrt{1/m}\right)$. 
\end{example}


\section{Conclusion} 
\label{sec:conclusion}
In this paper, we consider weighted ERM of the symmetric \problemabbr{} problem and establish new universal risk bounds under the ``low-noise'' condition. 
These new bounds are tighter in the case of incomplete graphs and can be degenerate to the known tightest bound when graphs are complete. 
Based on this result, one can train a classifier with a better risk bound by putting proper weights on training examples. 
We propose an efficient algorithm to obtain the approximate optimal weighting vector and prove that the algorithm is a FPTAS for the weighting vector optimization problem. 
Finally, we discuss two cases to show the merits of our new risk bounds.


\section*{Acknowledgments}

The authors would like to thank the anonymous reviewers for their helpful comments. This work was supported by Guangdong Shannon Intelligent Tech.\ co., Ltd., National Key R$\&$D Program of China (2017YFC0803700), National Natural Science Foundation of China (61502320, 61420106013, 61521092) and Science Foundation of Shenzhen City in China (JCYJ20160419152942010).


\bibliography{Mendeley}
\bibliographystyle{aaai}

\section*{Appendix} 
\appendix
\label{sec:appendix}

This appendix is organized as follows. Section \ref{sec:fractional_coloring_approach} 
provides new risk bounds for weighted ERM following the Janson's decomposition and show that it cannot improve prior results. 
Section \ref{sec:weighted_risk_bounds} establishes universal risk bounds for weighted ERM on i.i.d. examples under the "low-noise" condition and proves upper bounds involved in covering number for weighted empirical processes. 
Section \ref{sec:a_moment_inequality_for_weighted_u_processes} presents the classical symmetrization and randomization tricks and decoupling inequality for degenerated weighted $U$-processes and the degenerated part $\widetilde{U}_\weight{}(r)$.
Then, some useful inequalities including the moment inequality are proved for weighted Rademacher chaos.
We mainly use these inequalities to bound $U_\weight{}(r)$ and $\widetilde{U}_\weight{}(r)$.
Section \ref{sec:technical_proofs} provides technical proofs omitted from the main track and the appendix.
For the sake of completeness, we present the Khinchine inequality and the metric entropy inequality for Rademacher chaos in Section \ref{sub:metric_entropy_inequality}.

\section{Janson's Decomposition} 
\label{sec:fractional_coloring_approach}

There are literatures that use Janson's decomposition derived from \cite{janson2004large} to study the problem learning from networked examples \cite{Usunier2005,Biau2006,ralaivola2009chromatic,DBLP:conf/icml/RalaivolaA15}.
They usually model the networked data with the line graph of $G$.
\begin{definition}[line graph]
  \label{de:line graph}
  Let $G=(V,E)$ be a data graph we consider in the main track. We define the line graph of $G$ as a graph $D_G = (D_V, D_E)$, in which $D_V = E$ and $\edge{i,j}\in D_E$ if and only if $e_i\cap e_j\neq\emptyset$ in $G$.
\end{definition}
This framework differs from our setting and detains less information from the data graph, as argued in Section~\ref{sub:intiutions}. 
One can analyze this framework by the fractional coloring and the Janson's decomposition.

\begin{definition}[fractional coloring]
  Let $D_G=(D_V,D_E)$ be a graph. $\mathcal{C}=\set{(\mathcal{C}_j,q_j)}_{j\in (1,\dots,J)}$, for some positive integer $J$, with $\mathcal{C}_j\subseteq D_V$ and $q_j\in [0,1]$ is an fractional coloring of $D_G$, if
  \begin{itemize}
    \item $\forall j$, $\mathcal{C}_j$ is an independent set, i.e., there is no connection between vertices in $C_j$.
    \item it is an exact cover of $G$: $\forall v \in D_V$, $\sum_{j: v\in \mathcal{C}_j} q_j = 1$.
  \end{itemize}
\end{definition}
The weight $W(\mathcal{C})$ of $\mathcal{C}$ is given by $W(\mathcal{C})=\sum_{j=1}^J q_j$ and the minimum weight $\fcoloring{}(D_G)=\min_{\mathcal{C}} W(\mathcal{C})$ over the set of all fractional colorings is the \emph{fractional chromatic number} of $D_G$.
By the Janson's decomposition, one splits all examples into several independent sets according to a fractional coloring of $D_G$ and then analyze each set using the standard method for i.i.d.\ examples.

In particular, Theorem~\ref{th:fractional_mcdiarmid} \citep{also see}{Theorem~$2$}{Usunier2005} provides a variation of McDiarmid's theorem using the Janson's decomposition.
For simplicity, let $\trainingset=\set{z_i}_{i=1}^{m}$ be the training examples drawn from $\zspace^m$ and $\trainingset{}_{\mathcal{C}_j}$ be the examples included in $\mathcal{C}_j$.
Also, let $k_j^i$ be the index of the $i$-th example of $\mathcal{C}_j$ in the training set $\trainingset{}$.

\begin{theorem}
   \label{th:fractional_mcdiarmid}
   Using the notations defined above, let $\mathcal{C} = \set{(\mathcal{C}_j, q_j)}_{j=1}^J$ be a fractional coloring of $D_G$. Let $f:\zspace^m \to \real{}$ such that
   \begin{itemize}
     \item there exist $J$ functions $\zspace^{|\mathcal{C}_j|} \to \real{}$ which satisfy $\forall S\in \zspace^m$, $f(S)=\sum_{j=1}^J q_j f_j(S_{\mathcal{C}_j})$.
     \item there exist $c_i,\dots,c_m\in \real{}_+$ such that $\forall j,\forall S_{\mathcal{C}_j}, S_{\mathcal{C}_j}^k$ such that $\trainingset{}_{\mathcal{C}_j}$ and $\trainingset{}_{\mathcal{C}_j}^k$ differ only in the $k$-th example, $|f_j(S_{\mathcal{C}_j}) - f_j(S_{\mathcal{C}_j}^k)|\le c_{k_j^i}$.
   \end{itemize}
     Then, we have
     \[\Pro[f(S)-\E[f(S)]\ge \epsilon] \le \exp\left(-\frac{2\epsilon^2}{\fcoloring{}(D_G)\sum_{i=1}^m c_i^2}\right).\]
 \end{theorem}

\subsection{Weighted ERM} 
\label{sub:weighted_erm}

Using the above theorem, we show that the risk bounds for weighted ERM derived from the Janson's decomposition cannot improve the results of \cite{Usunier2005}, as shown in the following theorem.

\begin{theorem}
\label{th:fractional_coloring_weighted}
  Consider a minimizer $r_\weight{}$ of the weighted empirical risk $\empiricalrisk{\weight{}}$ over a class $R$. For all $\delta\in (0,1]$, with probability at least $1-\delta$, we have
  \[\risk{}(r)-\empiricalrisk{\weight{}}(r)\le \rademachercomplexity_{\weight{}}^*(R,S)+\sqrt{\fcoloring(D_G)\log(1/\delta)}\frac{\|\weight{}\|_2}{\normo{\weight{}}}\]
  where
  \[\rademachercomplexity_{\weight{}}^*(R, S) = \frac{2}{N}\E_\rademacher{}\left[\sum_{j=1}^J q_j\sup_{r\in R} \sum_{i\in \mathcal{C}_j}w_{k_j^i}\rademacher{}_i r(z_{k_j^i})\right].\]
  is the weighted empirical fractional Rademacher complexity of $R$ with respect to $D_G$.
\end{theorem}

In this setting, although the weights $\weight{}$ are no limit to the fractional matching, the risk bounds cannot improve the results of \cite{Usunier2005}, which is of the order $\sqrt{\fcoloring{}(D_G)/m}$, as $\|\weight\|_2/\normo{\weight{}}\le 1/\sqrt{m}$.

\subsection{``Low-noise'' Condition} 
\label{sub:low_noise_condition}

If we considering the complete graph with equal weighting scheme, by the Janson's decomposition, the empirical risk $\empiricalrisk{m}(L)$ can be represented as an average of sums of i.i.d.\ r.v.'s
\SingleColumn
\begin{equation}
  \label{eq:fractional_coloring_ERM}
  \frac{1}{m!}\sum_{\bm{\pi}\in\mathcal{G}_n}\frac{1}{\lfloor m/2\rfloor}\sum_{i=1}^{\lfloor m/2 \rfloor} \ell(r, (X_{\pi(i)}, X_{\pi(i)+\lfloor m/2\rfloor}, Y_{\pi(i), \pi(i)+\lfloor m/2\rfloor}))
\end{equation}
\SingleColumnEnd
\DoubleColumn
\begin{equation}
\begin{aligned}
  \label{eq:fractional_coloring_ERM}
  \frac{1}{m!}\sum_{\bm{\pi}\in\mathcal{G}_n}\frac{1}{\lfloor m/2\rfloor}\sum_{i=1}^{\lfloor m/2 \rfloor} \ell(r, (X_{\pi(i)}, X_{\pi(i)+\lfloor m/2\rfloor}, \\
  Y_{\pi(i), \pi(i)+\lfloor m/2\rfloor}))
  \end{aligned}
\end{equation}
\DoubleColumnEnd
where the sum is taken over all permutations of $\mathcal{G}_m$, the symmetric group of order $m$, and $\lfloor u \rfloor$ denotes the integer part of any $u\in \real{}$.
From \cite{Biau2006}, the bounds for excess risk of \eqref{eq:fractional_coloring_ERM} are of the order $O(1/\sqrt{n})$.
Moreover, by the result in \cite{DBLP:conf/icml/RalaivolaA15}, tighter risk bounds may be obtained under the following assumption which can lead to ``low-noise'' condition \cite{tsybakov2004optimal,Massart2006}.
\begin{assumption}
  There exists $C>0$ and $\theta\in [0,1]$ such that for all $\epsilon>0$,
  \[\Pro\left[|\eta(X_1,X_2)-\frac{1}{2}|\le \epsilon\right]\le C\epsilon^{\theta/(1-\theta)}.\]
\end{assumption}
The risk bounds of the order $O(\log(n)/n)$ may be achieved if $\theta=1$ \cite{Massart2006}, which however is very restrictive.
We can use the example in \citet{papa2016graph} to show this.
\begin{example}
Let $N$ be a positive integer. For each vertex $i \in \set{1,\dots,n}$, we observe $X_i = (X_i^1, X_i^2)$ where $X_i^1$ and $X_i^2$ are two distinct elements drawn from $\set{1,\dots,N}$.
This may, for instance, correspond to the two preferred items of a user $i$ among a list of $N$ items.
Consider now the case that two nodes are likely to be connected if they share common preference, e.g., $Y_{i,j} \sim Ber(\#(X_i\cap X_j)/2)$. One can easily check that $\Pro[|\eta(X_1,X_2)-1/2|=0]>0$, so tight risk bounds cannot be obtained for minimizers of \eqref{eq:fractional_coloring_ERM}.
\end{example}

\section{Universal Risk Bounds for Weighted ERM on i.i.d.\ Examples} 
\label{sec:weighted_risk_bounds}

In section 3, the excess risk is split into two types of processes: the weighted empirical process of i.i.d.\ examples and two degenerated processes.
In this section, we prove general risk bounds for the weighted empirical process of i.i.d.\ examples. 
The main idea is similar to \cite{Massart2006} that tighter bounds for the excess risk can be obtained if the variance of the excess risk is controlled by its expected value.

\subsection{Bennett Concentration Inequality} 
\label{subsub:bennett_type_inequality}


First, we prove a concentration inequality for the supremum of weighted empirical processes derived from \cite{Bousquet2002a}.

\begin{theorem}
    \label{th:weighted_bennett_inequality}
    Assume the $(X_1,\dots,X_i)$ are i.i.d.\ random variable according to $P$. Let $\mathcal F$ be a countable set of functions from $\mathcal X$ to $\mathbb R$ and assume that all functions $f$ in $\mathcal F$ are $P$-measurable and square-integrable. If $\sup_{f\in\mathcal F} |f| \le b$, we denote
    \[Z = \sup_{f\in \mathcal F} \frac{1}{\normo{\weight{}}}\sum_{i=1}^n w_i (f(X_i)-\E[f(X_i)]).\]
    which $\{w_i\}_{i=1}^n$ are bounded weights such that $0\le w_i\le 1$ for $i=1,\dots,n$ and $\normo{\weight{}}>0$.
    Let $\sigma$ be a positive real number such that $\sigma^2\ge \sup_{f\in\mathcal F} \Var[f(X)]$ almost surely, then for all $x\ge 0$, we have
    \begin{equation}
        \label{eq:weighted_bennett_inequality}
            \Pro\left[Z-\E[Z] \ge \sqrt{\frac{2(\frac{\|\weight{}\|_2^2}{\normo{\weight{}}}\sigma^2+4b\E[Z])x}{\normo{\weight{}}}} + \frac{2bx}{3\normo{\weight{}}}\right] \le e^{-x}.
    \end{equation}
\end{theorem}

It is a variant of Theorem 2.3 of \cite{Bousquet2002a} by just applying Theorem 2.1 of \cite{Bousquet2002a} with the weighted empirical process.
Then, by Theorem \ref{th:weighted_bennett_inequality}, we can generalize the results of \cite{Massart2006} to weighted ERM. We start by describing the probabilistic framework adapts to our problem.

\subsection{General Upper Bounds} 
\label{sub:i_i_d_}


Suppose that one observes independent variable $\xi_1,\dots,\xi_n$ taking their values in some measurable space $\mathcal Z$ with common distribution $P$. 
For every $i$, the variable $\xi_i=(X_i,Y_i)$ is a copy of a pair of random variables $(X,Y)$ where $X$ take its values in measurable space $\xspace$. 
Think of $\mathcal{R}$ as being the set of all measurable functions from $\xspace$ to $\{0,1\}$. 
Then we consider some \emph{loss} function
\begin{equation}
    \label{eq:iid_loss}
    \gamma: \mathcal{R} \times \mathcal{Z} \to [0,1].
\end{equation}
Basically one can consider some set $\mathcal{R}$, which is known to contain the \emph{Bayes classifier} $r^*$ that achieves the best (smallest) \emph{expected loss} $P[\gamma(r,\cdot)]$ when $r$ varies in $\mathcal{R}$. The \emph{relative expected loss $\relossf$} is defined by
\begin{equation}
    \label{eq:iid_relative_expected_loss}
    \relossf(r^*,r) = \Pro[\gamma(r,\cdot)-\gamma(r^*,\cdot)], \forall r\in\mathcal{R}
\end{equation}
Since the empirical process on i.i.d.\ examples split from the excess risk is with non-negative weights on all examples, we define the \emph{weighted loss} as
\begin{equation}
    \label{eq:weighted_erm}
    \gamma_\weight(r) = \frac{1}{\normo{\weight{}}}\sum_{i=1}^n w_i\gamma(r,\xi_i)
\end{equation}
where $0\le w_i\le 1$ for all $i=1,\dots,n$ and $\normo{\weight{}}=\sum_{i=1}^n w_i> 0$. Weighted ERM approach aims to find a minimizer of the weighted empirical loss $\hat{r}$ in the hypothesis set $R\subset \mathcal{R}$ to approximate $r^*$.

We introduce the \emph{weighted centered empirical process} $\cenprocess{\weight}$ defined by
\begin{equation}
    \label{eq:centered_empirical_process}
    \cenprocess{\weight}(r) = \gamma_\weight(r)-\Pro[\gamma(r,\cdot)].
\end{equation}
In addition to the relative expected loss function $\relossf$, we shall need another way to measure the closeness between the elements of $R$.
\LongVersion
which is directly connected to the variance of the increments of $\cenprocess{\weight}$ and therefore will play an important role in the analysis of the fluctuations of $\cenprocess{\weight}$.
\LongVersionEnd
Let $d$ be some pseudo-distance on $\mathcal{R}\times\mathcal{R}$ such that
\begin{equation}
    \label{eq:variancD_Eistance_inequality}
    \Var[\gamma(r,\cdot)-\gamma(r^*,\cdot)]\le d^2(r,r^*), \forall r\in\mathcal{R}.
\end{equation}

A tighter risk bound for weighted ERM is derived from Theorem \ref{th:weighted_bennett_inequality} which combines two different moduli of uniform continuity: the stochastic modulus of uniform continuity of $\cenprocess{\weight}$ over $R$ with respect to $d$ and the modulus of uniform continuity of $d$ with respect to $\relossf$.

Next, we need to specify some mild regularity conditions functions that we shall assume to be verified by the moduli of continuity involved in the following result.

\begin{definition}
\label{con:nondecreasing_condition}
    We denote by $D$ the class of nondecreasing and continuous functions $\psi$ from $\mathbb R_+$ to $\mathbb R_+$ such that $x\to \psi(x)/x$ is nonincreasing on $(0,+\infty)$ and $\psi(1)\ge 1$.
\end{definition}

In order to avoid measurability problems, we need to consider some separability condition on $R$. The following one will be convenient.

\begin{assumption}
\label{con:separability_condition}
    There exists some countable subset $R^\prime$ of $R$ such that, for every $r\in R$, there exists some sequence $\{r_k\}$ of elements of $R^\prime$ such that, for every $\xi \in \mathcal Z$, $\gamma(r_k, \xi)$ tends to $\gamma(r,\xi)$ as $k$ tends to infinity.
\end{assumption}

The upper bound for the relative expected loss of any empirical risk minimizer on some given model $R$ will depend on the bias term $\relossf(r^*,R)=\inf_{r\in R}\relossf(r^*,r)$ and the fluctuations of the empirical process $\cenprocess{\weight{}}$ on $R$. As a matter of fact, we shall consider some slightly more general estimators. Namely, given some nonnegative number $\rho$, we consider some $\rho$-empirical risk minimizer, that is, any estimator $r$ taking its values in $R$ such that $\gamma_\weight(\hat{r})\le \rho+\inf_{r\in R}\gamma_\weight(r)$.

\begin{theorem}[risk bound for weighted ERM]
    \label{th:weighted_erm_risk_bounds}
    Let $\gamma$ be a loss function such $r^*$ minimizes $\Pro[\gamma(r,\cdot)]$ when $r$ varies in $\mathcal{R}$. Let $\phi$ and $\psi$ belong to the class of functions $D$ defined above and let $R$ be a subset of $\mathcal R$ satisfying the separability Assumption \ref{con:separability_condition}. Assume that, on the one hand,
    \begin{equation}
        \label{eq:weighted_erm_risk_bound_con1}
        d(r^*,r)\le \frac{\sqrt{\normo{\weight{}}}}{\|\weight{}\|_2}\psi(\sqrt{\relossf(r^*,r)}), \forall r\in \mathcal R,
    \end{equation}
    and that, on the other hand, one has, for every $r\in R^\prime$,
    \begin{equation}
        \label{eq:weighted_erm_risk_bound_con2}
        \sqrt{\normo{\weight{}}}\E\left[\sup_{r'\in R^\prime,\frac{\|\weight{}\|_2}{\sqrt{\normo{\weight{}}}}d(r',r)\le \sigma}[\cenprocess{\weight}(r')-\cenprocess{\weight}(r)\right]\le \phi(\sigma)
    \end{equation}
    for every positive $\sigma$ such that $\phi(\sigma)\le \sqrt{\normo{\weight{}}}\sigma^2$, where $R^\prime$ is given by Assumption \ref{con:separability_condition}. Let $\epsilon_*$ be the unique positive solution of the equation
    \begin{equation}
        \label{eq:risk_bound_con3}
        \sqrt{\normo{\weight{}}}\epsilon_*^2=\phi(\psi(\epsilon_*)).
    \end{equation}
    Then there exists an absolute constant $K$ such that, for every $y\ge 1$, the following inequality holds:
    \begin{equation}
        \label{eq:risk_bound_probability}
        \Pro\left[\relossf(r^*,\hat{r})>2\rho+2\relossf(r^*,R)+K y\epsilon_*^2\right]\le e^{-y}.
    \end{equation}
\end{theorem}


\subsection{Maximal Inequality for Weighted Empirical Processes} 
\label{sub:maximal_inequality_for_weighted_processes}


Next, we present the maximal inequality involved in covering number for weighted empirical processes. Let us fix some notation. We consider i.i.d.\ random variables $\xi_1,\dots,\xi_n$ with values in some measurable space $\mathcal{Z}$ and common distribution $P$. For any $P$-integrable function $f$ on $\mathcal{Z}$, we define $P_\weight(f)=\frac{1}{\normo{\weight{}}}w_i\sum_{i=1}^nf(\xi_i)$ and $v_\weight(f)=P_\weight(f)-P(f)$ where $0\le w_i\le 1$ for all $i=1,\dots,n$ and $\normo{\weight{}}=\sum_{i=1}^n w_i> 0$. Given a collection $\mathcal{F}$ of $P$-integrable functions $f$, our purpose is to control the expectation of $\sup_{f\in\mathcal{F}}v_\weight(f)$ or $\sup_{f\in\mathcal{F}}-v_\weight(f)$.

\begin{lemma}
\label{le:maximal_entropy_inequality}
    Let $\mathcal{F}$ be a countable collection of measurable functions such that $f\in [0,1]$ for every $f\in\mathcal{F}$, and let $f_0$ be a measurable function such that $f_0\in[0,1]$. Let $\sigma$ be a positive number such that $\Pro[|f-f_0|]\le \sigma^2$ for every $f\in\mathcal{F}$. Then, setting
    \[\varphi(\sigma)=\int_0^{\sigma} (\log N_\infty(\mathcal{F},\epsilon^2))^{1/2}d\epsilon,\]
    the following inequality is available:
    \DoubleColumn
    \begin{align*}
        \sqrt{\normo{\weight{}}}\max(\E[\sup_{f\in\mathcal{F}}v_\weight(f_0-f)],\E[\sup_{f\in\mathcal{F}} &v_\weight(f-f_0)])\\
        &\le 12\varphi(\sigma).
    \end{align*}
    \DoubleColumnEnd
    \SingleColumn
    \[\sqrt{\normo{\weight{}}}\max\left(\E[\sup_{f\in\mathcal{F}}v_\weight(f_0-f)],\E[\sup_{f\in\mathcal{F}} v_\weight(f-f_0)]\right)\le 12\varphi(\sigma).\]
    \SingleColumnEnd
    provided that $4\varphi(\sigma)\le \sigma^2\sqrt{\normo{\weight{}}}$.
\end{lemma}


\section{Inequalities for $U_\weight{}(r)$ and $\widetilde{U}_\weight{}(r)$} 
\label{sec:a_moment_inequality_for_weighted_u_processes}

In this section, we first show the classical symmetrization and randomization tricks for the degenerated weighted $U$-statistics $U_\weight{}(r)$ and the degenerated part $\widetilde{U}_\weight{}(r)$.
Then we establish general exponential inequalities for weighted Rademacher chaos.
This result is generalized from \cite{clemenccon2008ranking} based on moment inequalities obtained for empirical processes and Rademacher chaos in \cite{Boucheron2005}.
With this moment inequality, we prove the inequality for weighted Rademacher chaos, which involves the $\lnorm{}_\infty$ covering number of the hypothesis set.

\begin{lemma}[decoupling and undecoupling]
    \label{le:weighted_u_statistics_decoupling}
    Let $(X_i^\prime)_{i=1}^n$ be an independent copy of the sequence $(X_i)_{i=1}^n$. Then, for all $q\ge 1$, we have:
    \DoubleColumn
    \begin{equation}
        \begin{aligned}
        \label{eq:weighted_u_statistics_decoupling}
            \E[\sup_{f_{i,j}\in\mathcal{F}}&|\sum_{\pair{i,j}\in E}w_{i,j} f_{i,j}(X_i,X_j)|^q] \\
            &\le 4^q\E[\sup_{f_{i,j}\in\mathcal{F}}|\sum_{\pair{i,j}\in E} w_{i,j}f_{i,j}(X_i,X_j^\prime)|^q]
        \end{aligned}
    \end{equation}
    \DoubleColumnEnd
    \SingleColumn
    \begin{equation}
        \begin{aligned}
        \label{eq:weighted_u_statistics_decoupling}
            \E[\sup_{f_{i,j}\in\mathcal{F}}&|\sum_{\pair{i,j}\in E}w_{i,j} f_{i,j}(X_i,X_j)|^q] \le 4^q\E[\sup_{f_{i,j}\in\mathcal{F}}|\sum_{\pair{i,j}\in E} w_{i,j}f_{i,j}(X_i,X_j^\prime)|^q]
        \end{aligned}
    \end{equation}
    \SingleColumnEnd
    If the functions $f_{i,j}$ are symmetric in the sense that for all $X_i,X_j$,
    \[f_{i,j}(X_i,X_j)=f_{j,i}(X_j,X_i)\]
    and $(w_{i,j})_{\pair{i,j}\in E}$ is symmetric, then the inequality can be reversed, that is,
    \DoubleColumn
    \begin{equation}
        \begin{aligned}
        \label{eq:weighted_u_statistic_coupling}
        \E[\sup_{f_{i,j}\in\mathcal{F}}&|\sum_{\pair{i,j}\in\E}w_{i,j} f_{i,j}(X_i,X_j^\prime)|^q]\\
        &\le 4^q \E[\sup_{f_{i,j}\in\mathcal{F}}|\sum_{\pair{i,j}\in E}w_{i,j} f_{i,j}(X_i,X_j)|^q]
        \end{aligned}
    \end{equation}
    \DoubleColumnEnd
    \SingleColumn
    \begin{equation}
        \begin{aligned}
        \label{eq:weighted_u_statistic_coupling}
        \E[\sup_{f_{i,j}\in\mathcal{F}}&|\sum_{\pair{i,j}\in\E}w_{i,j} f_{i,j}(X_i,X_j^\prime)|^q] \le 4^q \E[\sup_{f_{i,j}\in\mathcal{F}}|\sum_{\pair{i,j}\in E}w_{i,j} f_{i,j}(X_i,X_j)|^q]
        \end{aligned}
    \end{equation}
    \SingleColumnEnd
\end{lemma}
\begin{lemma}[randomization]
    \label{le:weighted_u_statistics_randomization}
    Let $(\sigma_i)_{i=1}^n$ and $(\sigma_i^\prime)_{i=1}^n$ be two independent sequences of i.i.d.\ Radermacher variables, independent from the $(X_i,X_i^\prime)$'s. If $f$ is degenerated, we have for all $q\ge 1$,
    \DoubleColumn
    \begin{align*}
        \E[\sup_{f\in\mathcal{F}}&|\sum_{\pair{i,j}\in E} w_{i,j}f(X_i,X_j^\prime)|^q]\\
        &\le 4^q\E[\sup_{f\in\mathcal{F}}|\sum_{\pair{i,j}\in E}\sigma_i\sigma_j^\prime w_{i,j}f(X_i, X_j^\prime)|^q]
    \end{align*}
    \DoubleColumnEnd
    \SingleColumn
    \begin{align*}
        \E[\sup_{f\in\mathcal{F}}&|\sum_{\pair{i,j}\in E} w_{i,j}f(X_i,X_j^\prime)|^q]\le 4^q\E[\sup_{f\in\mathcal{F}}|\sum_{\pair{i,j}\in E}\sigma_i\sigma_j^\prime w_{i,j}f(X_i, X_j^\prime)|^q]
    \end{align*}
    \SingleColumnEnd
\end{lemma}
\begin{lemma}
    \label{le:weighted_decoupling_undecoupling}
    Let $(X_i^\prime)_{i=1}^n$ be an independent copy of the sequence $(X_i)_{i=1}^n$. Consider random variables valued in $\{0,1\}$, $(\tilde{Y}_{i,j})_{\pair{i,j}\in E}$, conditionally independent given the $X_i^\prime$'s and the $X_i$'s and such that $\Pro[\tilde{Y}_{i,j}=1\mid X_i,X_j^\prime]=\eta(X_i, X_j^\prime)$. We have for all $q\ge 1$,
    \DoubleColumn
    \begin{equation}
        \begin{aligned}
        \label{eq:weighted_u_statistics_decoupling}
            \E[\sup_{r\in R}&|\sum_{\pair{i,j}\in E}w_{i,j} \tilde{h}_r(X_i,X_j, Y_{i,j})|^q] \\
            &\le 4^q\E[\sup_{r\in R}|\sum_{\pair{i,j}\in E} w_{i,j}\tilde{h}_r(X_i,X_j^\prime, \tilde{Y}_{i,j})|^q]
        \end{aligned}
    \end{equation}
    \DoubleColumnEnd
    \SingleColumn
    \begin{equation}
        \begin{aligned}
        \label{eq:weighted_u_statistics_decoupling}
            \E[\sup_{r\in R}&|\sum_{\pair{i,j}\in E}w_{i,j} \tilde{h}_r(X_i,X_j, Y_{i,j})|^q] \le 4^q\E[\sup_{r\in R}|\sum_{\pair{i,j}\in E} w_{i,j}\tilde{h}_r(X_i,X_j^\prime, \tilde{Y}_{i,j})|^q]
        \end{aligned}
    \end{equation}
    \SingleColumnEnd
    and the inequality can be reversed,
    \DoubleColumn
    \begin{equation}
        \begin{aligned}
        \label{eq:weighted_u_statistic_coupling}
        \E[\sup_{r\in R}&|\sum_{\pair{i,j}\in\E}w_{i,j} \tilde{h}_r(X_i,X_j^\prime, \tilde{Y}_{i,j})|^q]\\
        &\le 4^q \E[\sup_{r\in R}|\sum_{\pair{i,j}\in E}w_{i,j} \tilde{h}_r(X_i,X_j,Y_{i,j})|^q]
        \end{aligned}
    \end{equation}
    \DoubleColumnEnd
    \SingleColumn
    \begin{equation}
        \begin{aligned}
        \label{eq:weighted_u_statistic_coupling}
        \E[\sup_{r\in R}&|\sum_{\pair{i,j}\in\E}w_{i,j} \tilde{h}_r(X_i,X_j^\prime, \tilde{Y}_{i,j})|^q] \le 4^q \E[\sup_{r\in R}|\sum_{\pair{i,j}\in E}w_{i,j} \tilde{h}_r(X_i,X_j,Y_{i,j})|^q]
        \end{aligned}
    \end{equation}
    \SingleColumnEnd
\end{lemma}

\begin{lemma}
    \label{le:weighted_randomization}
    Let $(\sigma_i)_{i=1}^n$ and $(\sigma_i^\prime)_{i=1}^n$ be two independent sequences of i.i.d.\ Radermacher variables, independent from the $(X_i,X_i^\prime, Y_{i,j}, \tilde{Y}_{i,j})$'s. Then, we have for all $q\ge 1$,
    \DoubleColumn
    \begin{align*}
        \E[\sup_{r\in R}&|\sum_{\pair{i,j}\in E} w_{i,j}\tilde{h}_r(X_i,X_j^\prime, \tilde{Y}_{i,j})|^q]\\
        &\le 4^q\E[\sup_{r\in R}|\sum_{\pair{i,j}\in E}\sigma_i\sigma_j^\prime w_{i,j}\tilde{h}_r(X_i, X_j^\prime,\tilde{Y}_{i,j})|^q]
    \end{align*}
    \DoubleColumnEnd
    \SingleColumn
    \begin{align*}
        \E[\sup_{r\in R}&|\sum_{\pair{i,j}\in E} w_{i,j}\tilde{h}_r(X_i,X_j^\prime, \tilde{Y}_{i,j})|^q] \le 4^q\E[\sup_{r\in R}|\sum_{\pair{i,j}\in E}\sigma_i\sigma_j^\prime w_{i,j}\tilde{h}_r(X_i, X_j^\prime,\tilde{Y}_{i,j})|^q]
    \end{align*}
    \SingleColumnEnd
\end{lemma}

Lemma \ref{le:weighted_u_statistics_decoupling} and \ref{le:weighted_u_statistics_randomization} are applications of Theorem 3.1.1 and Theorem 3.5.3 of \cite{de2012decoupling} respectively, providing decoupling and randomization inequalities for degenerated weighted $U$-statistics of order $2$. Lemma \ref{le:weighted_decoupling_undecoupling} and \ref{le:weighted_randomization} are the variants of Lemma \ref{le:weighted_u_statistics_decoupling} and \ref{le:weighted_u_statistics_randomization} respectively, which is suitable for the degenerated part $\widetilde{U}_\weight{}(r)$.

\begin{theorem}[moment inequality]
    \label{th:moment_inequality_of_weighted_u_statistics}
    Let $X, X_1,\dots,X_n$ be i.i.d.\ random variables and let $\mathcal F$ be a class of kernels. Consider a weighted Rademacher chaos $Z_\rademacher{}$ of order $2$ on the graph $G=(V,E)$ indexed by $\mathcal F$,
    \[Z = \sup_{f\in\mathcal F}|\sum_{\pair{i,j}\in E}w_{i,j}f(X_i,X_j)|\]
    where $\E[f(X,x)]=0$ for all $x\in\xspace{},f\in\mathcal{F}$. Assume also for all $x,x'\in\xspace$, $f(x,x')=f(x',x)$ (symmetric) and $\sup_{f\in\mathcal F}\|f\|_\infty=F$. Let $(\rademacher{}_i)_{i=1}^n$ be i.i.d.\ Rademacher random variables and introduce the random variables
    \[Z_\rademacher = \sup_{f\in\mathcal F}|\sum_{\pair{i,j}\in E}w_{i,j}\rademacher_i\rademacher_jf(X_i,X_j)|\]
    \[U_\rademacher = \sup_{f\in\mathcal F}\sup_{\alpha:\|\alpha\|_2\le 1}\sum_{\pair{i,j}\in\E}w_{i,j}\rademacher_i\alpha_jf(X_i,X_j)\]
    \[M=\sup_{f\in\mathcal F,k=1,\dots,n}|\sum_{i:(i,k)\in E}w_{i,k}\rademacher_if(X_i,X_k)|\]
    Then exists a universal constant $C$ such that for all $n$ and $t>0$,
    \DoubleColumn
    \begin{align*}
        \Pro[Z&\ge C\E[Z_\rademacher]+t]\\
        &\le \exp\bigg(-\frac{1}{C}\min\Big((\frac{t}{\E[U_\rademacher]})^2,\frac{t}{\E[M]+F\|\weight{}\|_2},\\
        &\qquad(\frac{t}{\|\weight{}\|_{\max}F})^{2/3},\sqrt{\frac{t}{\|\weight{}\|_\infty F}}\Big)\bigg)
    \end{align*}
    \DoubleColumnEnd
    \SingleColumn
    \begin{align*}
        \Pro[Z&\ge C\E[Z_\rademacher]+t]\\
        &\le \exp\bigg(-\frac{1}{C}\min\Big((\frac{t}{\E[U_\rademacher]})^2,\frac{t}{\E[M]+F\|\weight{}\|_2},(\frac{t}{\|\weight{}\|_{\max}F})^{2/3},\sqrt{\frac{t}{\|\weight{}\|_\infty F}}\Big)\bigg)
    \end{align*}
    \SingleColumnEnd
    where $\|\weight{}\|_{\max}=\max_i \sqrt{\sum_{j:\pair{i,j}\in E} w_{i,j}^2}$.
\end{theorem}

If the hypothesis set $\mathcal{F}$ is a subset of $\lnormspace{}_\infty(\xspace^2)$ (upper bounds on the uniform covering number with $\lnorm{}_\infty$ metric can be calculated \cite{cucker2007learning}), we show $\E[Z_\rademacher{}]$, $\E[U_\rademacher{}]$ and $\E[M]$ can be bounded by $N_\infty(\mathcal{F},\epsilon)$ since all these Rademacher random variables satisfy the Khinchine inequality (see Section~\ref{sub:metric_entropy_inequality}). Following the metric entropy inequality for Khinchine-type processes (see Section~\ref{sub:metric_entropy_inequality}), it is easy to get the following Corollary.

\begin{corollary}
    \label{cor:covering_number_inequality}
    With the same setting of Theorem~\ref{th:moment_inequality_of_weighted_u_statistics}, if $\mathcal{F}\subset \lnormspace{}_\infty(\xspace{}^2)$, we have for any $\delta<1/e$,
    \[\Pro[Z\le \kappa]\ge 1-\delta\]
    where 
    \SingleColumn
    \begin{align*}
        \kappa\le&C\bigg(\|\weight{}\|_2\int_0^{2F}\log N_\infty(\mathcal F, \epsilon)d\epsilon+\max\Big(\|\weight{}\|_2\log(1/\delta)\int_0^{2F}\sqrt{\log N_\infty(\mathcal F,\epsilon)}d\epsilon,\\
        &\qquad\qquad(\log(1/\delta))^{3/2}\|\weight{}\|_{\max}, (\log(1/\delta))^2\|\weight{}\|_\infty\Big)\bigg).
    \end{align*}
    \SingleColumnEnd
    \DoubleColumn
    \begin{align*}
        \kappa\le&C\bigg(\|\weight{}\|_2\int_0^{2F}\log N_\infty(\mathcal F, \epsilon)d\epsilon\\
        &+\max\Big(\|\weight{}\|_2\log(1/\delta)\int_0^{2F}\sqrt{\log N_\infty(\mathcal F,\epsilon)}d\epsilon,\\
        &\qquad(\log(1/\delta))^{3/2}\|\weight{}\|_{\max}, (\log(1/\delta))^2\|\weight{}\|_\infty\Big)\bigg).
    \end{align*}
    \DoubleColumnEnd
    with a universal constant $C$.
\end{corollary}

\section{Technical Proofs} 
\label{sec:technical_proofs}

\subsection{Proofs Omitted in Section~\ref{sec:risk_bounds}} 
\label{sub:proofs_omitted_in_section_sec:risk_bounds}

\begin{proof}[Proof of Lemma~\ref{le:uniform_approximation}] 
Since $R\subset\lnormspace{}_\infty(\xspace{}^2)$ (Assumption~\ref{ass:covering_number}), by Corollary~\ref{cor:covering_number_inequality}, the weighted degenerated $U$-process $\sup_{r\in R}|U_\weight{}(r)|$ can be bounded by the $\lnorm{}_\infty$ covering number of $R$, that is, for any $\delta\in (0,1/e)$, we have
\[\Pro[\sup_{r\in R}|U_\weight{}(r)|\le \kappa]\ge 1-\delta\]
where 
\SingleColumn
\begin{align*}
    \kappa\le&\frac{C_1}{\normo{\weight}}\bigg(\|\weight{}\|_2\int_0^{1}\log N_\infty(\mathcal F, \epsilon)d\epsilon+\max\Big(\|\weight{}\|_2\log(1/\delta)\int_0^{1}\sqrt{\log N_\infty(\mathcal F,\epsilon)}d\epsilon,\\
    &\qquad\qquad(\log(1/\delta))^{3/2}\|\weight{}\|_{\max}, (\log(1/\delta))^2\|\weight{}\|_\infty\Big)\bigg).
\end{align*}
\SingleColumnEnd
\DoubleColumn
\begin{align*}
    \kappa\le&\frac{C_1}{\normo{\weight}}\bigg(\|\weight{}\|_2\int_0^{1}\log N_\infty(R, \epsilon)d\epsilon\\
    &+\max\Big(\|\weight{}\|_2\log(1/\delta)\int_0^{1}\sqrt{\log N_\infty(R,\epsilon)}d\epsilon,\\
    &\qquad(\log(1/\delta))^{3/2}\|\weight{}\|_{\max},(\log(1/\delta))^2\|\weight{}\|_\infty\Big)\bigg).
\end{align*}
\DoubleColumnEnd
with a universal constant $C_1<\infty$. 
Then the first inequality for $\sup_{r\in R}|U_\weight{}(r)|$ follows the fact that $R$ satisfies Assumption~\ref{ass:covering_number}.
Similarly, by Lemma~\ref{le:weighted_decoupling_undecoupling} and Lemma~\ref{le:weighted_randomization}, we can convert the moment of $\sup_{r\in R}|\widetilde{U}_\weight{}(r)|$ to the moment of Rademacher chaos
\[4^q\E[\sup_{r\in R}|\sum_{\pair{i,j}\in E}\sigma_i\sigma_j^\prime w_{i,j}\tilde{h}_r(X_i, X_j^\prime,\tilde{Y}_{i,j})|^q]\] 
which can be handled by the by-products of Theorem~\ref{th:moment_inequality_of_weighted_u_statistics}. More specifically, using \eqref{eq:inequality_rademacher_chaos_1} and \eqref{eq:inequality_rademacher_chaos_2} combined with the arguments in Corollary~\ref{cor:covering_number_inequality} and Assumption~\ref{ass:covering_number} will gives the second inequality for $\sup_{r\in R}|\widetilde{U}_\weight{}(r)|$.
\end{proof}


\begin{proof}[Proof of Lemma~\ref{le:variant_control}] 
    For any function $r\in\mathcal{R}$, observe first that
    \DoubleColumn
    \begin{align*}
        \E[&q_r(X_1,X_2,Y_{1,2})\mid X_1]\\
        &= \E[\E[q_r(X_1,X_2,Y_{1,2})\mid X_1,X_2]\mid X_1]\\
        &= \E[|1-2\eta(X_1,X_2)|\indicator{}_{r(X_1,X_2)\neq r^*(X_1,X_2)}\mid X_1]
    \end{align*}
    \DoubleColumnEnd
    \SingleColumn
    \begin{align*}
        \E[q_r(X_1,X_2,Y_{1,2})\mid X_1]&= \E[\E[q_r(X_1,X_2,Y_{1,2})\mid X_1,X_2]\mid X_1]\\
        &= \E[|1-2\eta(X_1,X_2)|\indicator{}_{r(X_1,X_2)\neq r^*(X_1,X_2)}\mid X_1]
    \end{align*}
    \SingleColumnEnd
    Then observing that
    \[|1-2\eta(X_1,X_2)|^2\le |1-2\eta(X_1,X_2)|\]
    almost sure, and combining with Jensen inequality, we have
    \SingleColumn
    \begin{align*}
        \Var[\E[q_r(X_1,X_2,Y_{1,2})\mid X_1]] & \le \E[(\E[q_r(X_1,X_2,Y_{1,2})\mid X_1])^2]\\
        &\le \E[|1-2\eta(X_1,X_2)|\indicator{}_{r(X_1,X_2)\neq r^*(X_1,X_2)}]\\
        &= \Lambda(r).
    \end{align*}
    \SingleColumnEnd
    \DoubleColumn
    \begin{align*}
        \Var&[\E[q_r(X_1,X_2,Y_{1,2})\mid X_1]] \\
        & \le \E[(\E[q_r(X_1,X_2,Y_{1,2})\mid X_1])^2]\\
        &\le \E[|1-2\eta(X_1,X_2)|\indicator{}_{r(X_1,X_2)\neq r^*(X_1,X_2)}]\\
        &= \Lambda(r).
    \end{align*}
    \DoubleColumnEnd
\end{proof}

\begin{proof}[Proof of Lemma~\ref{le:risk_bounds_iid}] 
    First, we introduce some notations of weighted ERM of the i.i.d.\ case. 
    We denote $\{\bar{w}_i=\sum_{j:\pair{i,j}\in E}w_{i,j}: i=1,\dots,n\}$ the weights on vertices and introduce the ``loss function''
    \[\gamma(r, X)=2h_r(X)+\Lambda{}(r)\]
    and the weighted empirical loss of vertices
    \[\gamma_\verticeweight{}(r)=\frac{1}{\normo{\verticeweight{}}}\sum_{i=1}^n\bar{w}_i\gamma(r,X_i)=T_\weight{}(r).\]
    Define centered empirical process
    \[\cenprocess{\verticeweight}(r)=\frac{1}{\normo{\verticeweight{}}}\sum_{i=1}^n\bar{w}_i(\gamma(r,X_i)-\Lambda(r))\]
    and the pseudo-distance
    \[d(r,r^\prime) = \frac{\sqrt{\normo{\verticeweight{}}}}{\|\verticeweight\|_2}\left(\E[(\gamma(r,X)-\gamma(r',X))^2]\right)^{1/2}\]
    for every $r,r^\prime\in R$.
    Let $\phi$ be
    \begin{equation}
        \label{eq:proof_phi}
        \phi(\sigma)=12\int_0^{\sigma}(\log N_\infty(R,\epsilon^2))^{1/2}d\epsilon.
    \end{equation}
    From the definition of ``loss function'' $\gamma$, we have the excess risk of $r$ is
    \[\relossf(r,r^*)=\Lambda(r)-\Lambda(r^*)=\Lambda(r).\]
    According to Lemma~\ref{le:uniform_approximation}, as $\Lambda^2(r)\le \Lambda(r)$, we have for every $r\in R$,
    \begin{align*}
        d(r,r^*)\le \frac{\sqrt{\normo{\verticeweight{}}}}{\|\verticeweight\|_2}\sqrt{5\relossf(r,r^*)}
    \end{align*}
    which implies that the modulus of continuity $\psi$ can be taken as
    \begin{equation}
        \label{eq:proof_psi}
        \psi(\epsilon)=\sqrt{5}\epsilon.
    \end{equation}
    Then from Lemma~\ref{le:maximal_entropy_inequality}, we have
    \[\sqrt{\normo{\verticeweight{}}}\E\left[\sup_{r^\prime\in R,\frac{\|\verticeweight\|_2}{\sqrt{\normo{\verticeweight{}}}}d(r,r^\prime)\le \sigma}|\cenprocess{\verticeweight}(r)-\cenprocess{\verticeweight}(r^\prime)|\right]\le \phi(\sigma).\]
    provided that $\phi(\sigma)/3\le \sqrt{\normo{\verticeweight{}}}\sigma^2$.
    It remains to bound the excess risk of $r_\weight{}$ by the tight bounds for weighted ERM on i.i.d.\ examples by Theorem~\ref{th:weighted_erm_risk_bounds}. For any $\delta\in(0,1)$, we have with probability at least $1-\delta$,
    \DoubleColumn
    \begin{equation}
        \begin{aligned}
            \label{eq:main_result_proof_1}
            \risk(r_\weight{})-\bayeserror \le 2(&\inf_{r\in R}\risk(r)-\bayeserror) + 2\rho \\
            &+ K\log(1/\delta)\epsilon_*^2)
        \end{aligned}
    \end{equation}
    \DoubleColumnEnd
    \SingleColumn
    \begin{equation}
        \begin{aligned}
            \label{eq:main_result_proof_1}
            \risk(r_\weight{})-\bayeserror \le 2(\inf_{r\in R}\risk(r)-\bayeserror) + 2\rho + C\log(1/\delta)\epsilon_*^2)
        \end{aligned}
    \end{equation}
    \SingleColumnEnd
    where $C$ is a universal constant and $\epsilon_*$ is the unique positive solution of the equation
    \[\sqrt{\normo{\verticeweight{}}}\epsilon_*^2=\phi(\psi(\epsilon_*)).\]
    When $ R$ satisfies Assumption~\ref{ass:covering_number}, there exists a universal constant $C^\prime$ such that
    \[\epsilon_*^2\le C^\prime K^{1/(1+\complexbound{})}(\frac{1}{(1-\complexbound{})^2\normo{\verticeweight{}}})^{1/(1+\complexbound{})}\]
    which completes the proof.
\end{proof}


\subsection{Proof Omitted in Section \ref{sec:fractional_coloring_approach}} 
\label{sub:proofs_omitted_in_section_a}
\begin{proof}[Proof of Theorem~\ref{th:fractional_coloring_weighted}]
We write, for all $r\in R$,
\SingleColumn
\begin{equation}
  \label{eq:fractional_coloring_decomposition}
    \begin{aligned}
    \risk{}(r)-\empiricalrisk{\weight{}}(r)&\le \sup_{r\in R}\left[\E[\ell(r,z)] - \frac{1}{\normo{\weight{}}}\sum_{i=1}^m w_i\ell(r,z_i)]\right]\\
    &\le \sum_{j=1}^J p_j\left[ \sup_{r\in R}\sum_{i\in\mathcal{C}_j} \frac{w_{k_j^i}}{\normo{\weight{}}} (\E[\ell(r,z)] - \ell(r,z_{k_j^i}))\right].
    \end{aligned}
\end{equation}
\SingleColumnEnd
\DoubleColumn
\begin{equation}
  \label{eq:fractional_coloring_decomposition}
    \begin{aligned}
    \risk{}(r)-\empiricalrisk{\weight{}}(r)&\le \sup_{r\in R}\left[\E[\ell(r,z)] - \frac{1}{\normo{\weight{}}}\sum_{i=1}^m w_i\ell(r,z_i)]\right]\\
    &\le \sum_{j=1}^J p_j\left[ \sup_{r\in R}\sum_{i\in\mathcal{C}_j} \frac{w_{k_j^i}}{\normo{\weight{}}} (\E[\ell(r,z)] - \ell(r,z_{k_j^i}))\right].
    \end{aligned}
\end{equation}
\DoubleColumnEnd
Now, consider, for each $j$,
\[f_j(S_{\mathcal{C}_j}) = \sup_{r\in R}\sum_{i\in\mathcal{C}_j} \frac{w_{k_j^i}}{\normo{\weight{}}} (\E[\ell(r,z)] - \ell(r,z_{k_j^i})).\]
Let $f$ is defined by, for all training set $\trainingset{}$, $f(\trainingset{})=\sum_{j=1}^Jp_jf_j(S_{\mathcal{C}_j})$, then $f$ satisfies the conditions of Theorem~\ref{th:fractional_mcdiarmid} with, for any $i\in \set{1,\dots,m}$, $\beta_i\le w_i/\normo{\weight{}}$. Therefore, we can claim that, with probability at least $1-\delta$,
\SingleColumn
\begin{align*}
  \risk{}(r) - \empiricalrisk{\weight{}}(r)&\le \E\left[\sup_{r\in R}\left[\E[\ell(r,z)] - \frac{1}{\normo{\weight{}}}\sum_{i=1}^m w_i\ell(r,z_i)]\right]\right] + \sqrt{\fcoloring(D_G)\log(1/\delta)}\frac{\|\weight{}\|_2}{\normo{\weight{}}}
\end{align*}
\SingleColumnEnd
\DoubleColumn
\begin{align*}
  \risk{}(r) - \empiricalrisk{\weight{}}(r)&\le \E\left[\sup_{r\in R}\left[\E[\ell(r,z)] - \frac{1}{\normo{\weight{}}}\sum_{i=1}^m w_i\ell(r,z_i)]\right]\right]\\
  &\qquad+ \sqrt{\fcoloring(D_G)\log(1/\delta)}\frac{\|\weight{}\|_2}{\normo{\weight{}}}
\end{align*}
\DoubleColumnEnd
Then, using the standard symmetrization technique \citep{see}{Theorem~$4$}{Usunier2005}, one can bound the first item in the right hand side by $\mathfrak{R}_{\weight{}}^*(R,S)$ which completes the proof. 
\end{proof}

\subsection{Proofs Omitted in Section \ref{sec:weighted_risk_bounds}} 
\label{sec:proof_omitted_in_section_sec:weighted_risk_bounds}


\begin{proof}[Proof of Theorem \ref{th:weighted_bennett_inequality}] 
    We use an auxiliary random variable
    \[\widetilde{Z}=\frac{\normo{\weight{}}}{2b}Z=\sup_{f\in\mathcal{F}}\frac{1}{2b}\sum_{i=1}^n w_i\allowbreak(f(X_i)-\E[f(X_i)]).\]
    We denote by $f_k$ a function such that
    \[f_k = \frac{1}{2b}\sup_{f\in\mathcal F} \sum_{i\neq k}w_i(f(X_i)-\E[f(X_i)]).\]
    We introduce following auxiliary random variables for $k=1,\dots,n$,
    \[Z_k = \frac{1}{2b}\sup_{f\in\mathcal F} \sum_{i\neq k}w_i(f(X_i-\E[f(X_i)]))\]
    and
    \[Z_k^\prime = \frac{1}{2b}w_i(f(X_k)-\E[f(X_k)]).\]
    Denoting by $f_0$ the function achieving the maximum in $Z$, we have
    \[\widetilde{Z}-Z_k\le \frac{1}{2b}w_i(f_0(X_k)-\E[f_0(X_k)])\le 1\ a.s.,\]
    \[\widetilde{Z}-Z_k-Z_k^\prime\ge 0\]
    and
    \[\E[Z_k^\prime] = 0.\]
    The first inequality is derived from $w_i\le 1$ and $\sup_{f\in\mathcal{F}, X\in\mathcal{X}} f(X)-\E[f(X)]\le 2b$.
    Also, we have
    \DoubleColumn
    \begin{align*}
        (n-1)\widetilde{Z} &= \sum_{k=1}^n\frac{1}{2b}\sum_{i\neq k}w_i(f_0(X_i)-\E[f_0(X_i)])\\
        &\le \sum_{k=1}^n Z_k,
    \end{align*}
    \DoubleColumnEnd
    \SingleColumn
    \begin{align*}
        (n-1)\widetilde{Z} &= \sum_{k=1}^n\frac{1}{2b}\sum_{i\neq k}w_i(f_0(X_i)-\E[f_0(X_i)])\\
        &\le \sum_{k=1}^n Z_k,
    \end{align*}
    \SingleColumnEnd
    and
    \begin{align*}
        \sum_{k=1}^n \E_n^k[Z_k^{\prime2}] &= \frac{1}{2b}\sum_{k=1}^n \E[w_i^2(f_k(X_k)-\E[f_k(X_k)])^2]\\
        &\le \frac{1}{4b^2}\|\weight{}\|_2^2\sup_{f\in\mathcal F}\Var[f(X)]\\
        &\le \frac{1}{4b^2}\|\weight{}\|_2^2\sigma^2.
    \end{align*}
    where $\sigma^2\ge \sup_{f\in\mathcal F}\Var[f(X)]$.
    Notice that we use the fact the $X_i$ have identical distribution.
    Applying Theorem 1 of \cite{Bousquet2002a} with $v=2\E[\widetilde{Z}]+\frac{\|\weight{}\|_2^2}{4b^2}\sigma^2$ will give
    \[\Pro[\widetilde{Z}-\E[\widetilde{Z}]\ge \sqrt{2vx}+\frac{x}{3}]\le e^{-x},\]
    and then
    \DoubleColumn
    \begin{align*}
        \Pro[\frac{\normo{\weight{}}}{2b}(Z-\E[Z])&\ge \sqrt{2x(\frac{\normo{\weight{}}}{b}\E[Z]+\frac{\|\weight{}\|_2^2}{4b^2}\sigma^2)}\\
        &\qquad+\frac{x}{3}] \le e^{-x}
    \end{align*}
    \DoubleColumnEnd
    \SingleColumn
    \begin{align*}
        \Pro\left[\frac{\normo{\weight{}}}{2b}(Z-\E[Z])\ge \sqrt{2x(\frac{\normo{\weight{}}}{b}\E[Z]+\frac{\|\weight{}\|_2^2}{4b^2}\sigma^2)}+\frac{x}{3}\right] \le e^{-x}
    \end{align*}
    \SingleColumnEnd
    which proves the inequality.
\end{proof}


\begin{proof}[Proof of Theorem \ref{th:weighted_erm_risk_bounds}] 
    Since $R$ satisfies Condition \ref{con:separability_condition}, we notice that, by dominated convergence, for every $r\in R$, considering the sequence $\{r_k\}$ provided by Condition \ref{con:separability_condition}, one has $\Pro[\gamma(\cdot,r_k)]$ that tends to $\Pro[\gamma(\cdot, r)]$ as $k$ tends to infinity. Denote the bias term of loss $\relossf(r^*,R)=\inf_{r\in R}\relossf(r^*,r)$. Hence, $\relossf(r^*,R)=\relossf(r^*,R^\prime)$, which implies that there exists some point $\pi(r^*)$ (which may depend on $\epsilon_*$) such that $\pi(r^*)\in R^\prime$ and
    \begin{equation}
        \label{eq:asymptotic_function_inequality}
        \relossf(r^*, \pi(r^*))\le \relossf(r^*,R)+\epsilon_*^2.
    \end{equation}

    We start from the identity
    \SingleColumn
    \[\relossf(r^*,\hat{r})=\relossf(r^*,\pi(r^*))+\gamma_\weight(\hat{r})-\gamma_\weight(\pi(r^*))+\cenprocess{\weight}(\pi(r^*))-\cenprocess{\weight}(\hat{r})\]
    \SingleColumnEnd
    \DoubleColumn
    \begin{align*}
        \relossf(r^*,\hat{r})&=\relossf(r^*,\pi(r^*))+\gamma_\weight(\hat{r})-\gamma_\weight(\pi(r^*))\\
        &\qquad+\cenprocess{\weight}(\pi(r^*))-\cenprocess{\weight}(\hat{r})
    \end{align*}
    \DoubleColumnEnd
    which, by definition of $\hat{r}$, implies that
    \[\relossf(r^*,\hat{r})\le \rho+\relossf(r,\pi(r^*))+\cenprocess{\weight}(\pi(r^*))-\cenprocess{\weight}(\hat{r}).\]
    Let $x=\sqrt{K^\prime y}\epsilon_*$, where $K^\prime$ is a constant to be chosen later such that $K^\prime\ge 1$ and
    \[V_x=\sup_{r\in R}\frac{\cenprocess{\weight}(\pi(r^*))-\cenprocess{\weight}(r)}{\relossf(r^*,r)+\epsilon^*+x^2}.\]
    Then,
    \[\relossf(r^*,\hat{r})\le \rho+\relossf(r^*,\pi(r^*))+V_x(\relossf(r^*,\hat{r})+x^2+\epsilon_*^2)\]
    and therefore, on the event $V_x<1/2$, one has
    \[\relossf(r^*,\hat{r})\le 2(\rho+\relossf(r^*,\pi(r^*)))+\epsilon_*^2+x^2,\]
    yielding
    \SingleColumn
    \begin{equation}
        \label{eq:risk_bounds_convert}
        \Pro[\relossf(r^*,\hat{r})\le 2(\rho+\relossf(r^*,\pi(r^*)))+3\epsilon_*^2+x^2]\le \Pro[V_x\ge \frac{1}{2}].
    \end{equation}
    \SingleColumnEnd
    \DoubleColumn
    \begin{equation}
        \begin{aligned}
            \label{eq:risk_bounds_convert}
            \Pro[\relossf(r^*,\hat{r})\le 2(\rho+\relossf(r^*,\pi(r^*)))+3\epsilon_*^2+x^2]\\
            \le \Pro[V_x\ge \frac{1}{2}].
        \end{aligned}
    \end{equation}
    \DoubleColumnEnd
    Since $\relossf$ is bounded by 1, we may always assume $x$ (and thus $\epsilon_*$) to be not larger than 1. Assuming that $x\le 1$, it remains to control the variable $V_x$ via Theorem \ref{th:weighted_bennett_inequality}. In order to use Theorem \ref{th:weighted_bennett_inequality}, we first remark that, by Condition \ref{con:separability_condition},
    \[V_x=\sup_{r\in R^\prime}\frac{\cenprocess{\weight}(\pi(r^*))-\cenprocess{\weight}(r)}{\relossf(r^*,r)+\epsilon^*+x^2}\]
    which means that we indeed have to deal with a countably indexed empirical process. Note that the triangle inequality implies via \eqref{eq:variancD_Eistance_inequality}, \eqref{eq:asymptotic_function_inequality} and \eqref{eq:weighted_erm_risk_bound_con1} that
    \begin{equation}
        \begin{aligned}
            \label{eq:bound_of_variance}
            (\Var[\gamma(r,\cdot)-\gamma(\pi(r^*),\cdot)])^{\frac{1}{2}}&\le d(r^*,r)+d(r^*, \pi(r^*))\\
            &\le 2\frac{\sqrt{\normo{\weight{}}}}{\|\weight{}\|_2}\psi(\sqrt{\relossf(r^*,r)+\epsilon_*^2})
        \end{aligned}
    \end{equation}
    Since $\gamma$ takes its values in $[0,1]$, introducing the functions $\psi_1=\min(1,2\psi)$ and, we derive from \eqref{eq:bound_of_variance} that
    \DoubleColumn
    \begin{align*}
        \sup_{r\in R}\Var[\frac{\cenprocess{\weight}(\pi(r^*))-\cenprocess{\weight}(t)}{\relossf(r^*,t)+\epsilon^*+x^2}]&\le \sup_{\epsilon\ge 0}\frac{(\frac{\sqrt{\normo{\weight{}}}}{\|\weight{}\|_2}\psi_1(\epsilon))^2}{(\epsilon^2+x^2)^2}\\
        &\le \frac{\normo{\weight{}}}{\|\weight{}\|_2^2x^2}\sup_{\epsilon\ge 0}(\frac{\psi_1(\epsilon)}{\max(\epsilon,x)})^2.
    \end{align*}
    \DoubleColumnEnd
    \SingleColumn
    \[\sup_{r\in R}\Var[\frac{\cenprocess{\weight}(\pi(r^*))-\cenprocess{\weight}(r)}{\relossf(r^*,r)+\epsilon^*+x^2}]\le \frac{\normo{\weight{}}}{\|\weight{}\|_2^2}\sup_{\epsilon\ge 0}\frac{\psi_1^2(\epsilon)}{(\epsilon^2+x^2)^2}\le \frac{\normo{\weight{}}}{\|\weight{}\|_2^2x^2}\sup_{\epsilon\ge 0}(\frac{\psi_1(\epsilon)}{\max(\epsilon,x)})^2.\]
    \SingleColumnEnd
    Now the monotonicity assumptions on $\psi$ imply that either $\psi(\epsilon)\le \psi(x)$ if $x\ge \epsilon$ or $\psi(\epsilon)/\epsilon\le \psi(x)/x$ if $x\le \epsilon$. Hence, one has in any case $\psi(\epsilon)/(\max(\epsilon,x))\le \psi(x)/x$, which finally yields
    \[\sup_{r\in R}\Var[\frac{\cenprocess{\weight}(\pi(r^*))-\cenprocess{\weight}(r)}{\relossf(r^*,r)+\epsilon^*+x^2}]\le\frac{\normo{\weight{}}\psi_1^2(x)}{\|\weight{}\|_2^2x^4}.\]
    On the other hand, since $\gamma$ takes its values in $[0,1]$, we have
    \[\sup_{r\in R}\|\frac{\gamma(r,\cdot)-\gamma(\pi(r^*),\cdot)}{\relossf(r^*,r)+x^2}\|_\infty\le \frac{1}{x^2}.\]
    We can therefore apply Theorem~\ref{th:weighted_bennett_inequality} with $v=\psi_1^2(x)x^{-4}$ and $b=x^{-2}$, which gives that, on a set $\Omega_y$ with probability larger than $1-\exp(-y)$, the inequality
    \begin{equation}
        \begin{aligned}
            \label{eq:v_x_bennett_inequality}
            V_x < \E[V_x]+\sqrt{\frac{2y(\psi_1^2(x)x^{-2}+4\E[V_x])}{\normo{\weight{}}x^2}}+\frac{2y}{3\normo{\weight{}}x^2}.
        \end{aligned}
    \end{equation}
    Now since $\epsilon_*$ is assumed to be not larger than 1, one has $\psi(\epsilon_*)\ge \epsilon_*$ and therefore, for every $\sigma\ge \psi(\epsilon_*)$, the following inequality derives from the definition of $\epsilon_*$ by monotonicity:
    \[\frac{\phi(\sigma)}{\sigma^2}\le \frac{\phi(\psi(\epsilon_*))}{w^2(\epsilon_*)}\le \frac{\phi(\psi(\epsilon_*))}{\epsilon_*^2}=\sqrt{\normo{\weight{}}}.\]
    Thus, \eqref{eq:weighted_erm_risk_bound_con2} holds for every $\sigma\ge \psi(\epsilon_*)$. In order to control $\E[V_x]$, we intend to use Lemma A.5 of \cite{Massart2006}. 
    For every $r\in R^\prime$, we introduce $a^2(r)=\max(\relossf(r^*,\pi(r^*)),\relossf(r^*,r))$. Then by \eqref{eq:asymptotic_function_inequality}, $\relossf(r^*,r)\le a^2(r)\le \relossf(r^*,r)+\epsilon_*^2$. Hence, we have, on the one hand, that
    \[\E[V_x]\le \E[\sup_{r\in R^\prime}\frac{\cenprocess{\weight}(\pi(r^*))-\cenprocess{\weight}(r)}{a^2(r)+x^2}].\]
    and, on the other hand, that, for every $\epsilon\ge \epsilon_*$,
    \SingleColumn
    \[\E[\sup_{r\in R^\prime,a(r)\le \epsilon}\cenprocess{\weight}(\pi(r^*))-\cenprocess{\weight}(r)]\le \E[\sup_{r\in R^\prime,\relossf(r^*,r)\le \epsilon^2}\cenprocess{\weight}(\pi(r^*))-\cenprocess{\weight}(r)].\]
    \SingleColumnEnd
    \DoubleColumn
    \begin{align*}
        \E[\sup_{r\in R^\prime,a(r)\le \epsilon}&\cenprocess{\weight}(\pi(r^*))-\cenprocess{\weight}(r)]\\
        &\le \E[\sup_{r\in R^\prime,\relossf(r^*,r)\le \epsilon^2}\cenprocess{\weight}(\pi(r^*))-\cenprocess{\weight}(r)].
    \end{align*}
    \DoubleColumnEnd
    Now by \eqref{eq:asymptotic_function_inequality} if there exists some $r\in R^\prime$ such that $\relossf(r^*,r)\le \epsilon^2$, then $\relossf(r^*,\pi(r^*))\le \epsilon^2+\epsilon_*^2\le 2\epsilon^2$ and therefore, by assumption \eqref{eq:weighted_erm_risk_bound_con1} and monotonicity of $\theta\to \psi(\theta)/\theta$, $d(\pi(r^*),r)\le 2\frac{\sqrt{\normo{\weight{}}}}{\|\weight{}\|_2}\psi(\sqrt{2}\epsilon)\le 2\sqrt{2}\frac{\sqrt{\normo{\weight{}}}}{\|\weight{}\|_2}\psi(\epsilon)$, then $\frac{\|\weight{}\|_2}{\sqrt{\normo{\weight{}}}}d(\pi(r^*),r)\le 2\sqrt{2}\psi(\epsilon)$. Thus, we derive from \eqref{eq:weighted_erm_risk_bound_con2} that, for every $\epsilon\ge \epsilon_*$,
    \[\E[\sup_{r\in R^\prime,\relossf(r^*,r)\le \epsilon^2}\cenprocess{\weight}(\pi(r^*))-\cenprocess{\weight}(r)]\le \phi(2\sqrt{2}\psi(\epsilon))\]
    and since $\theta\to \phi(2\sqrt{2}\psi(\theta))/\theta$ is nonincreasing, we can use Lemma A.5 of \cite{Massart2006} to get
    \[\E[V_x]\le 4\phi(2\sqrt{2}\psi(x))/(\sqrt{\normo{\weight{}}}x^2),\]
    and by monotonicity of $\theta\to\phi(\theta)/\theta$,
    \[\E[V_x]\le 8\sqrt{2}\phi(\psi(x))/(\sqrt{\normo{\weight{}}}x^2).\]
    Thus, using the monotonicity of $\theta\to \phi(\psi(\theta))/\theta$, and the definition of $\epsilon_*$, we derive that
    \begin{equation}
        \label{eq:bound_of_expectation}
        \E[V_x]\le \frac{8\sqrt{2}\phi(\psi(\epsilon_*))}{\sqrt{\normo{\weight{}}}x\epsilon_*}=\frac{8\sqrt{2}\epsilon_*}{x}\le \frac{8\sqrt{2}}{\sqrt{K^\prime y}}\le \frac{8\sqrt{2}}{\sqrt{K^\prime}},
    \end{equation}
    provided that $x\ge \epsilon_*$, which holds since $K^\prime\ge 1$. Now, the monotonicity of $\theta\to\psi_1(\theta)/\theta$ implies that $x^{-2}\psi_1^2(x)\le \epsilon_*^{-2}\psi_1^2(\epsilon_*)$, but since $\phi(\theta)/\theta\ge \phi(1)\ge 1$ for every $\theta\in [0,1]$, we derive from \eqref{eq:risk_bound_con3} and the monotonicity of $\phi$ and $\theta\to\phi(\theta)/\theta$ that
    \[\frac{\psi_1^2(\epsilon_*)}{\epsilon_*^2}\le \frac{\phi^2(\psi_1(\epsilon_*))}{\epsilon_*^2}\le \frac{\phi^2(2\psi(\epsilon_*))}{\epsilon_*^2}\le 4\frac{\phi^2(\psi(\epsilon_*))}{\epsilon_*^2}\]
    and, therefore, $x^{-2}\psi_1^2(x)\le 4\normo{\weight{}}\epsilon_*^2$. Plugging this inequality together with \eqref{eq:bound_of_expectation} into \eqref{eq:v_x_bennett_inequality} implies that, on the set $\Omega_y$,
    \[V_x < \frac{8\sqrt{2}}{\sqrt{K^\prime}}+\sqrt{\frac{2y(4\normo{\weight{}}\epsilon_*^2+32/\sqrt{K^\prime})}{\normo{\weight{}}x^2}}+\frac{2y}{3\normo{\weight{}}x^2}.\]
    It remains to replace $x^2$ by its value $K^\prime y\epsilon_*^2$ to derive that, on the set $\Omega_y$, the following inequality holds:
    \[V_x < \frac{8\sqrt{2}}{\sqrt{K^\prime}}+\sqrt{\frac{8(1+4(\normo{\weight{}}\epsilon_*^2\sqrt{K^\prime})^{-1})}{K^\prime}}+\frac{2}{3\normo{\weight{}}K^\prime\epsilon_*^2}.\]
    Taking into account that $\phi(\psi(\theta))\ge \phi(\min(1,\psi(\theta)))\ge \theta$ for every $\theta\in [0,1]$, we deduce from the definition of $\epsilon_*$ that $\normo{\weight{}}\epsilon_*^2\ge 1$ and, therefore, the preceding inequality becomes, on $\Omega_y$,
    \[V_x < \frac{8\sqrt{2}}{\sqrt{K^\prime}}+\sqrt{\frac{8(1+4/\sqrt{K^\prime})}{K^\prime}}+\frac{2}{3K^\prime}.\]
    Hence, choosing $K^\prime$ as a large enough numerical constant guarantee that $V_x < 1/2$ on $\Omega_y$ and, therefore, \eqref{eq:risk_bounds_convert} yields
    \SingleColumn
    \begin{equation}
        \label{eq:risk_bounds_convert}
        \Pro[\relossf(r^*,\hat{r})\le 2(\rho+\relossf(r^*,\pi(r^*)))+3\epsilon_*^2+x^2]\le \Pro[\Omega_y^c]\le e^{-y}.
    \end{equation}
    \SingleColumnEnd
    \DoubleColumn
        \begin{align*}
            \label{eq:risk_bounds_convert}
            \Pro[\relossf(r^*,\hat{r})\le 2(\rho+\relossf(r^*,\pi(r^*)))&+3\epsilon_*^2+x^2]\\
            &\le \Pro[\Omega_y^c]\\
            &\le e^{-y}.
        \end{align*}
    \DoubleColumnEnd
    We get the required probability bound \eqref{th:weighted_erm_risk_bounds} by setting $K=K^\prime+3$.
\end{proof}

\begin{proof}[Proof of Lemma \ref{le:maximal_entropy_inequality}] 
    We first perform the control of $\E[\sup_{f\in\mathcal{F}}v_\weight(f-f_0)]$. For simplicity, we denote $H_\infty(\mathcal{F}, \epsilon)=\log N_\infty(\mathcal{F},\epsilon)$. For any integer $j$, we set $\sigma_j=\sigma2^{-j}$ and $H_j=H_\infty(\mathcal{F},\sigma_j^2)$. By definition of $H_j=H_\infty(\mathcal{F},\sigma_j^2)$, for any integer $j\ge 1$, we can define a mapping $\Pi_j$ from $\mathcal{F}$ to some finite collection of functions such that
    \begin{equation}
        \label{eq:bounded_entropy_mapping}
        \log \#\{\Pi_j\mathcal{F}\}\le H_j
    \end{equation}
    and
    \begin{equation}
        \label{eq:bounded_entropy_condition}
        \Pi_jf\le f \text{ with }P(f-\Pi_jf)\le \sigma_j^2, \forall f\in\mathcal{F}.
    \end{equation}
    For $j=0$, we choose $\Pi_0$ to be identically equal to $f_0$. For this choice of $\Pi_0$, we still have
    \begin{equation}
        \label{eq:bounded_entropy_f0}
        P(|f-\Pi_0f|)=P[|f-f_0|]\le \sigma_0^2=\sigma
    \end{equation}
    for every $f\in\mathcal{F}$. Furthermore, since we may always assume that the extremities of the balls used to cover $\mathcal{F}$ take their values in $[0,1]$, we also have for every integer $j$ that
    \[0\le\Pi_jf\le1.\]
    Noticing that since $u\to H_\infty(\mathcal{F},u^2)$ is nonincreasing,
    \[H_1\le \sigma_1^{-2}\varphi^2(\sigma),\]
    and under the condition $4\varphi(\sigma)\le \sigma^2\sqrt{\normo{\weight{}}}$, one has $H_1\le \sigma_1^2\normo{\weight{}}$. Thus, since $j\to H_j\sigma_j^{-2}$ increases to infinity, the set $\{j\ge 0: H_j\le \sigma_j^2\normo{\weight{}}\}$ is a nonvoid interval of the form
    \[\{j\ge 0: H_j\le \sigma_j^2\normo{\weight{}}\}=[0,J],\]
    with $J\ge 1$. For every $f\in\mathcal{F}$, starting from the decomposition
    \DoubleColumn
    \begin{align*}
    -v_\weight(f)=\sum_{j=0}^{J-1}v_\weight(\Pi_jf)-v_\weight(\Pi_{j+1}f)\\
    +v_\weight(\Pi_Jf)-v_\weight(f),
    \end{align*}
    \DoubleColumnEnd
    \SingleColumn
    \begin{align*}
    -v_\weight(f)=\sum_{j=0}^{J-1}v_\weight(\Pi_jf)-v_\weight(\Pi_{j+1}f)+v_\weight(\Pi_Jf)-v_\weight(f),
    \end{align*}
    \SingleColumnEnd
    we derive, since $\Pi_J(f)\le f$ and $P(f-\Pi_j(f))\le \sigma_J^2$, that
    \[-v_\weight(f)=\sum_{j=0}^{J-1}v_\weight(\Pi_jf)-v_\weight(\Pi_{j+1}f)+\sigma_J^2\]
    and, therefore,
    \DoubleColumn
    \begin{equation}
        \begin{aligned}
            \label{eq:bounded_entropy_decomposition}
            \E[&\sup_{f\in\mathcal{F}}[-v_\weight(f)]]\\
            &\le \sum_{j=0}^{J-1}\E[\sup_{f\in\mathcal{F}}v_\weight(\Pi_jf)-v_\weight(\Pi_{j+1}f)]+\sigma_J^2.
        \end{aligned}
    \end{equation}
    \DoubleColumnEnd
    \SingleColumn
    \begin{equation}
        \begin{aligned}
            \label{eq:bounded_entropy_decomposition}
            \E[\sup_{f\in\mathcal{F}}[-v_\weight(f)]]\le \sum_{j=0}^{J-1}\E[\sup_{f\in\mathcal{F}}v_\weight(\Pi_jf)-v_\weight(\Pi_{j+1}f)]+\sigma_J^2.
        \end{aligned}
    \end{equation}
    \SingleColumnEnd
    Now, it follows from \eqref{eq:bounded_entropy_condition} and \eqref{eq:bounded_entropy_f0} that, for every integer $j$ and every $f\in\mathcal{F}$, one has
    \[P[|\Pi_jf-\Pi_{j+1}f|]\le \sigma_j^2+\sigma_{j+1}^2=5\sigma_{j+1}^2\]
    and, therefore, since $|\Pi_jf-\Pi_{j+1}f|\le 1$,
    \[P[|\Pi_jf-\Pi_{j+1}f|^2]\le 5\sigma_{j+1}^2.\]
    Moreover, \eqref{eq:bounded_entropy_mapping} ensures that the number of functions of the form $\Pi_jf-\Pi_{j+1}f$ when varies in $\mathcal{F}$ is not larger than $\exp(H_j+H_{j+1})\le \exp(2H_{j+1})$, Hence, we derive from the maximal inequality for random vectors \citep{see}{Lemma A.1}{Massart2006} and the by-product of the proof of Bernstein's inequality for the weighted sum of networked random variables \citep{see}{Lemma 16}{wang2017learning} that
    \DoubleColumn
    \begin{align*}
        \sqrt{\normo{\weight{}}}&\E[\sup_{f\in\mathcal{F}}[v_\weight(\Pi_jf)-v_\weight(\Pi_{j+1}f)]]\\
        &\le 2[\sigma_{j+1}\sqrt{5H_{j+1}}+\frac{1}{3\sqrt{\normo{\weight{}}}}H_{j+1}]\\
    \end{align*}
    \DoubleColumnEnd
    \SingleColumn
    \begin{align*}
        \sqrt{\normo{\weight{}}}\E[\sup_{f\in\mathcal{F}}[v_\weight(\Pi_jf)-v_\weight(\Pi_{j+1}f)]]\le 2[\sigma_{j+1}\sqrt{5H_{j+1}}+\frac{1}{3\sqrt{\normo{\weight{}}}}H_{j+1}]\\
    \end{align*}
    \SingleColumnEnd
    because $w_i\le 1, \forall i\in 1,\dots,n$,
    and \eqref{eq:bounded_entropy_decomposition} becomes
    \DoubleColumn
    \begin{equation}
    \begin{aligned}
        \label{eq:bounded_entropy_result}
        &\sqrt{\normo{\weight{}}}\E[\sup_{f\in\mathcal{F}}-v_\weight(f)]\\
        &\le 2\sum_{j=1}^J[\sigma_j\sqrt{5H_j}+\frac{1}{3\sqrt{\normo{\weight{}}}}H_j]+4\sqrt{\normo{\weight{}}}\sigma_{J+1}^2.
    \end{aligned}
    \end{equation}
    \DoubleColumnEnd
    \SingleColumn
    \begin{equation}
    \begin{aligned}
        \label{eq:bounded_entropy_result}
        \sqrt{\normo{\weight{}}}\E[\sup_{f\in\mathcal{F}}-v_\weight(f)]\le 2\sum_{j=1}^J[\sigma_j\sqrt{5H_j}+\frac{1}{3\sqrt{\normo{\weight{}}}}H_j]+4\sqrt{\normo{\weight{}}}\sigma_{J+1}^2.
    \end{aligned}
    \end{equation}
    \SingleColumnEnd
    It follows from the definition of $J$ that, on the one hand, for every $j\le J$,
    \[\frac{1}{3\sqrt{\normo{\weight{}}}}H_j\le \frac{1}{3}\sqrt{H_j}\]
    and, on the other hand,
    \[4\sqrt{\normo{\weight{}}}\sigma_{J+1}^2\le 4\sigma_{J+1}\sqrt{H_{j+1}}.\]
    Hence, plugging these inequalities in \eqref{eq:bounded_entropy_result} yields
    \[\sqrt{\normo{\weight{}}}\E[\sup_{f\in\mathcal{F}}-v_\weight(f)]\le 6\sum_{j=1}^{J+1}\sigma_j\sqrt{H_j},\]
    and the result follows. The control of $\E[\sup_{f\in\mathcal{F}}v_\weight(f-f_0)]$ can be performed analogously.
\end{proof}

\subsection{Proofs Omitted in Section \ref{sec:a_moment_inequality_for_weighted_u_processes}} 
\label{sec:proofs_omitted_in_section_sec:a_moment_inequality_for_weighted_u_processes}

\begin{proof}[Proof of Lemma \ref{le:weighted_decoupling_undecoupling}] 
    This Lemma is derived from Lemma \ref{le:weighted_u_statistics_decoupling}, thus we can follow the similar arguments that can be found in \cite{de2012decoupling}.

    For any random variable $X$, we denote by $\mathcal L(X)$ its distribution. We denote by $\Sigma$ (respectively $\Sigma^\prime$) the sigma-filed generated by $\{X_1,\dots,X_n\}$ (respectively $\{X_1^\prime,\dots,X_n^\prime\}$). Let $(Y_{i,j}^\prime)_{(i,j)\in E}$ be Bernoulli random variables such that $\Pro[Y_{i,j}'=1\mid \Sigma, \Sigma^\prime]=\eta(X_i^\prime, X_j^\prime)$. Let $(\sigma_i)_{i=1}^n$ be independent Rademacher variables and define:
    \begin{align*}
        Z_i = X_i \text{ if } \sigma_i = 1 \text{, and } X_i^\prime \text{ otherwise,}\\
        Z_i^\prime = X_i^\prime \text{ if } \sigma_i = 1 \text{, and } X_i^\prime \text{ otherwise.}
    \end{align*}
    Conditionally upon the $X_i$ and $X_i^\prime$, the random vector $(Z_i, Z_i^\prime)$ takes the values $(X_i, X_i^\prime)$ or $(X_i^\prime, X_i)$, each with probability 1/2. In particular, we have:
    \SingleColumn
    \begin{equation}
        \label{eq:weighted_u_statistics_decoupling_fact1}
        \mathcal L(X_1,\dots,X_n,X_1^\prime,\dots,X_n^\prime) = \mathcal L(Z_1,\dots,Z_n,Z_1^\prime,\dots,Z_n^\prime).
    \end{equation}
    \SingleColumnEnd
    \DoubleColumn
    \begin{equation}
        \begin{aligned}
            \label{eq:weighted_u_statistics_decoupling_fact1}
            \mathcal L(X_1,\dots,X_n,X_1^\prime,\dots,X_n^\prime) =\\ \mathcal L(Z_1,\dots,Z_n,Z_1^\prime,\dots,Z_n^\prime)
        \end{aligned}
    \end{equation}
    \DoubleColumnEnd
    and
    \begin{equation}
        \label{eq:weighted_u_statistics_decoupling_fact2}
        \mathcal L(X_1,\dots,X_n) = \mathcal L(Z_1,\dots,Z_n).
    \end{equation}
    Let $\set{\tilde{Y}_{i,j}}_{\pair{i,j}\in E}$ be Bernoulli random variables such that $\Pro[\tilde{Y}_{i,j}=1 \mid \Sigma,\Sigma'] = \eta(X_i, X_j')$ and define for $\pair{i,j}\in E$,
    \begin{equation*}
        \hat{Y}_{i,j}=\begin{cases}
            Y_{i,j}&\text{if } \sigma_i=1\text{ and } \sigma_j=-1\\
            Y_{i,j}^\prime&\text{if } \sigma_i=-1\text{ and } \sigma_j=1\\
            \tilde{Y}_{i,j}&\text{if } \sigma_i=1\text{ and } \sigma_j=1\\
            \tilde{Y}_{i,j}&\text{if } \sigma_i=-1\text{ and } \sigma_j=-1\\
        \end{cases}
    \end{equation*}
    Notice that for all $f$,
    \DoubleColumn
    \begin{align*}
        \E_\sigma[&\tilde{h}_r(Z_i,Z_j^\prime,\tilde{Y}_{i,j})]\\
        &=\frac{1}{4}(\tilde{h}_r(X_i,X_j, Y_{i,j})+\tilde{h}_r(X_i^\prime,X_j,\tilde{Y}_{i,j})\\
        &\qquad+\tilde{h}_r(X_i,X_j^\prime,\tilde{Y}_{i,j})+\tilde{h}_r(X_i^\prime,X_j^\prime,Y_{i,j}^\prime))    
    \end{align*}
    \DoubleColumnEnd
    \SingleColumn
    \begin{align*}
        \E_\sigma[\tilde{h}_r(Z_i,Z_j^\prime,\hat{Y}_{i,j})]=\frac{1}{4}(\tilde{h}_r(X_i,X_j, Y_{i,j})+\tilde{h}_r(X_i^\prime,X_j,\tilde{Y}_{i,j})+\tilde{h}_r(X_i,X_j^\prime,\tilde{Y}_{i,j})+\tilde{h}_r(X_i^\prime,X_j^\prime,Y_{i,j}^\prime))    
    \end{align*}
    \SingleColumnEnd
    where $\E_\sigma$ denotes the expectation taken with respect to $\{\sigma_i\}_{i=1}^n$. Moreover, using
    \[\E[\tilde{h}_r(X_i^\prime,X_j^\prime,Y_{i,j}^\prime)\mid \Sigma]=0\]
    and (degenerated)
    \begin{align*}
        \E[\tilde{h}_r(X_i,X_j^\prime,\tilde{Y}_{i,j})\mid \Sigma] = 0\\
         \E[\tilde{h}_r(X_i^\prime,X_j,\tilde{Y}_{i,j})\mid \Sigma] = 0
    \end{align*}
    we easily get
    \[\tilde{h}_r(X_i,X_j,Y_{i,j})=4\E[\tilde{h}_r(Z_i,Z_j^\prime,\hat{Y}_{i,j})\mid \Sigma]\]
    For all $q\ge 1$, we therefore have
    \DoubleColumn
    \begin{align*}
        \E[&\sup_{r\in R}|\sum_{\pair{i,j}\in E}w_{i,j}\tilde{h}_r(X_i,X_j,Y_{i,j})|^q]\\
        &=\E[\sup_{r\in R}|\sum_{\pair{i,j}\in E}4w_{i,j}\E[\tilde{h}_r(Z_i,Z_j^\prime,\tilde{Y}_{i,j})\mid\Sigma]|^q]\\
        &\le 4^q\E[\sup_{r\in R}|\sum_{\pair{i,j}\in E}w_{i,j}\tilde{h}_r(Z_i,Z_j^\prime,\tilde{Y}_{i,j})|^q]
    \end{align*}
    \DoubleColumnEnd
    \SingleColumn
    \begin{align*}
        \E[\sup_{r\in R}|\sum_{\pair{i,j}\in E}w_{i,j}\tilde{h}_r(X_i,X_j,Y_{i,j})|^q]&=\E[\sup_{r\in R}|\sum_{\pair{i,j}\in E}4w_{i,j}\E[\tilde{h}_r(Z_i,Z_j^\prime,\hat{Y}_{i,j})\mid\Sigma]|^q]\\
        &\le 4^q\E[\sup_{r\in R}|\sum_{\pair{i,j}\in E}w_{i,j}\tilde{h}_r(Z_i,Z_j^\prime,\hat{Y}_{i,j})|^q]
    \end{align*}
    \SingleColumnEnd
    derived from the facts that the supreme and $|x|^p (p\ge1)$ are convex functions and the Jansen inequality. According to \eqref{eq:weighted_u_statistics_decoupling_fact1} and the fact that the distribution of $\hat{Y}_{i,j}$ only depends on the realization $Z_i,Z_j^\prime$, i.e.\ $\Pro[\hat{Y}_{i,j}\mid Z_i,Z_j^\prime]=\eta(Z_i,Z_j^\prime)$, we obtain
    \DoubleColumn
    \begin{align*}
        4^q\E[&\sup_{r\in R}|\sum_{\pair{i,j}\in E}w_{i,j}\hat{h}_r(Z_i,Z_j^\prime,\tilde{Y}_{i,j})|^q]\\
        &= 4^q\E[\sup_{r\in R}|\sum_{\pair{i,j}\in E}w_{i,j}\tilde{h}_r(X_i,X_j^\prime,\tilde{Y}_{i,j})|^q]
    \end{align*}
    \DoubleColumnEnd
    \SingleColumn
    \begin{align*}
        4^q\E[\sup_{r\in R}|\sum_{\pair{i,j}\in E}w_{i,j}\tilde{h}_r(Z_i,Z_j^\prime,\hat{Y}_{i,j})|^q]= 4^q\E[\sup_{r\in R}|\sum_{\pair{i,j}\in E}w_{i,j}\tilde{h}_r(X_i,X_j^\prime,\tilde{Y}_{i,j})|^q]
    \end{align*}
    \SingleColumnEnd
    which concludes the proof of \eqref{eq:weighted_u_statistics_decoupling}.

    By the symmetry of $\tilde{h}_r$ in the sense that $\tilde{h}_r(X_i,X_j,Y_{i,j})=\tilde{h}_r(X_j,X_i,Y_{j,i})$, we have
    \DoubleColumn
    \begin{align*}
        \E[&\sup_{r\in R} |\sum_{\pair{i,j}\in E}w_{i,j}\tilde{h}_r(X_i,X_j^\prime)|^q]\\
        &=\E[\sup_{r\in R} |\frac{1}{2}\sum_{\pair{i,j}\in E}w_{i,j}(\tilde{h}_r(X_i,X_j^\prime,\tilde{Y}_{i,j})\\
        &\qquad+\tilde{h}_r(X_i^\prime,X_j,\tilde{Y}_{i,j}))|^q]\\
        &= \E[\sup_{r\in R} |\frac{1}{2}\sum_{\pair{i,j}\in E}w_{i,j}(\tilde{h}_r(X_i,X_j^\prime,\tilde{Y}_{i,j})\\
        &\qquad+\tilde{h}_r(X_i^\prime,X_j,\tilde{Y}_{i,j})+\tilde{h}_r(X_i,X_j,Y_{i,j})\\
        &\qquad+\tilde{h}_r(X_i^\prime,X_j^\prime,Y^\prime_{i,j})\\
        &\qquad-\frac{1}{2}\sum_{\pair{i,j}\in E}w_{i,j}\tilde{h}_r(X_i,X_j,Y_{i,j})\\
        &\qquad-\frac{1}{2}\sum_{\pair{i,j}\in E}w_{i,j}\tilde{h}_r(X_i^\prime,X_j^\prime,Y^\prime_{i,j})|^q]\\
        &(\text{Triangle's Inequality and the convexity of }\sup_{r\in R}|\cdot|^q)\\
        &\le \frac{1}{2}\E[\sup_{r\in R} |\sum_{\pair{i,j}\in E}w_{i,j}(\tilde{h}_r(X_i,X_j^\prime,\tilde{Y}_{i,j})\\
        &\qquad+\tilde{h}_r(X_i^\prime,X_j,\tilde{Y}_{i,j})+\tilde{h}_r(X_i,X_j,Y_{i,j})\\
        &\qquad+\tilde{h}_r(X_i^\prime,X_j^\prime,Y^\prime_{i,j})|^q]\\
        &\qquad+ \frac{1}{4}\E[\sup_{r\in R} |2\sum_{\pair{i,j}\in E}w_{i,j}\tilde{h}_r(X_i,X_j,Y_{i,j})|^q] \\
        &\qquad+ \frac{1}{4}\E[\sup_{r\in R} |2\sum_{\pair{i,j}\in E}w_{i,j}\tilde{h}_r(X_i^\prime,X_j^\prime,Y^\prime_{i,j})|^q]\\
        &= \frac{1}{2}\E[\sup_{r\in R} |4\sum_{\pair{i,j}\in E}w_{i,j}\E_\sigma[\tilde{h}_r(Z_i,Z_j,\tilde{Y}_{i,j})]|^q]\\
        &\qquad+ \frac{1}{2}\E[\sup_{r\in R} |2\sum_{\pair{i,j}\in E}w_{i,j}\tilde{h}_r(X_i,X_j,Y_{i,j})|^q]\\
        &(\text{Jansen's Inequality and the convexity of }\sup_{r\in R}|\cdot|^q)\\
        &\le \frac{1}{2}\E[\sup_{r\in R} |4\sum_{\pair{i,j}\in E}w_{i,j}\tilde{h}_r(Z_i,Z_j,\tilde{Y}_{i,j})|^q] \\
        &\qquad+ \frac{1}{2}\E[\sup_{r\in R} |2\sum_{\pair{i,j}\in E}w_{i,j}\tilde{h}_r(X_i,X_j,Y_{i,j})|^q]\\
        &(\text{According to }\eqref{eq:weighted_u_statistics_decoupling_fact2})\\
        &= \frac{1}{2}\E[\sup_{r\in R} |4\sum_{\pair{i,j}\in E}w_{i,j}\tilde{h}_r(X_i,X_j,Y_{i,j})|^q] \\
        &\qquad+ \frac{1}{2}\E[\sup_{r\in R} |2\sum_{\pair{i,j}\in E}w_{i,j}\tilde{h}_r(X_i,X_j,Y_{i,j})|^q]\\
        &= 4^q\E[\sup_{r\in R} |\sum_{\pair{i,j}\in E}w_{i,j}\tilde{h}_r(X_i,X_j,Y_{i,j})|^q]
    \end{align*}
    \DoubleColumnEnd
    \SingleColumn
    \begin{align*}
        \E[&\sup_{r\in R} |\sum_{\pair{i,j}\in E}w_{i,j}\tilde{h}_r(X_i,X_j^\prime)|^q]\\
        &=\E[\sup_{r\in R} |\frac{1}{2}\sum_{\pair{i,j}\in E}w_{i,j}(\tilde{h}_r(X_i,X_j^\prime,\tilde{Y}_{i,j})+\tilde{h}_r(X_i^\prime,X_j,\tilde{Y}_{i,j}))|^q]\\
        &= \E[\sup_{r\in R} |\frac{1}{2}\sum_{\pair{i,j}\in E}w_{i,j}(\tilde{h}_r(X_i,X_j^\prime,\tilde{Y}_{i,j})+\tilde{h}_r(X_i^\prime,X_j,\tilde{Y}_{i,j})+\tilde{h}_r(X_i,X_j,Y_{i,j})+\tilde{h}_r(X_i^\prime,X_j^\prime,Y^\prime_{i,j})\\
        &\qquad-\frac{1}{2}\sum_{\pair{i,j}\in E}w_{i,j}\tilde{h}_r(X_i,X_j,Y_{i,j})-\frac{1}{2}\sum_{\pair{i,j}\in E}w_{i,j}\tilde{h}_r(X_i^\prime,X_j^\prime,Y^\prime_{i,j})|^q]\\
        &(\text{Triangle's Inequality and the convexity of }\sup_{r\in R}|\cdot|^q)\\
        &\le \frac{1}{2}\E[\sup_{r\in R} |\sum_{\pair{i,j}\in E}w_{i,j}(\tilde{h}_r(X_i,X_j^\prime,\tilde{Y}_{i,j})+\tilde{h}_r(X_i^\prime,X_j,\tilde{Y}_{i,j})+\tilde{h}_r(X_i,X_j,Y_{i,j})+\tilde{h}_r(X_i^\prime,X_j^\prime,Y^\prime_{i,j})|^q]\\
        &\qquad+\frac{1}{4}\E[\sup_{r\in R} |2\sum_{\pair{i,j}\in E}w_{i,j}\tilde{h}_r(X_i,X_j,Y_{i,j})|^q] + \frac{1}{4}\E[\sup_{r\in R} |2\sum_{\pair{i,j}\in E}w_{i,j}\tilde{h}_r(X_i^\prime,X_j^\prime,Y^\prime_{i,j})|^q]\\
        &= \frac{1}{2}\E[\sup_{r\in R} |4\sum_{\pair{i,j}\in E}w_{i,j}\E_\sigma[\tilde{h}_r(Z_i,Z_j,\tilde{Y}_{i,j})]|^q]+ \frac{1}{2}\E[\sup_{r\in R} |2\sum_{\pair{i,j}\in E}w_{i,j}\tilde{h}_r(X_i,X_j,Y_{i,j})|^q]\\
        &(\text{Jansen's Inequality and the convexity of }\sup_{r\in R}|\cdot|^q)\\
        &\le \frac{1}{2}\E[\sup_{r\in R} |4\sum_{\pair{i,j}\in E}w_{i,j}\tilde{h}_r(Z_i,Z_j,\tilde{Y}_{i,j})|^q] + \frac{1}{2}\E[\sup_{r\in R} |2\sum_{\pair{i,j}\in E}w_{i,j}\tilde{h}_r(X_i,X_j,Y_{i,j})|^q]\\
        &(\text{According to }\eqref{eq:weighted_u_statistics_decoupling_fact2})\\
        &= \frac{1}{2}\E[\sup_{r\in R} |4\sum_{\pair{i,j}\in E}w_{i,j}\tilde{h}_r(X_i,X_j,Y_{i,j})|^q] + \frac{1}{2}\E[\sup_{r\in R} |2\sum_{\pair{i,j}\in E}w_{i,j}\tilde{h}_r(X_i,X_j,Y_{i,j})|^q]\\
        &\le \E[\sup_{r\in R} |4\sum_{\pair{i,j}\in E}w_{i,j}\tilde{h}_r(X_i,X_j,Y_{i,j})|^q]\\
        &= 4^q\E[\sup_{r\in R} |\sum_{\pair{i,j}\in E}w_{i,j}\tilde{h}_r(X_i,X_j,Y_{i,j})|^q]
    \end{align*}
    \SingleColumnEnd
    which \eqref{eq:weighted_u_statistic_coupling} follows.
\end{proof}

\begin{proof}[Proof of Lemma \ref{le:weighted_randomization}] 
    Re-using the notations used in the proof of Lemma \ref{le:weighted_u_statistics_decoupling}, we further introduce $(X_i^{\prime\prime})_{i=1}^n$, a copy of $(X_i^\prime)_{i=1}^n$, independent from $\Sigma$, $\Sigma^\prime$, and denote by $\Sigma^{\prime\prime}$ its sigma-field. Let $(\tilde{Y}_{i,j}^{\prime\prime})_{(i,j)\in E}$ Bernoulli random variables such that $\Pro[\tilde{Y}_{i,j}^{\prime\prime}=1\mid \Sigma,\Sigma^\prime,\Sigma^{\prime\prime}]=\eta(X_i,X_j^{\prime\prime})$. We now use classic randomization techniques and introduce our "ghost" sample:
    \DoubleColumn
\begin{align*}
    &\E[\sup_{r\in R}|\sum_{\pair{i,j}\in E} w_{i,j}\tilde{h}_r(X_i,X_j^\prime,\tilde{Y}_{i,j}^{\prime\prime})|^q]\\
    &(\tilde{h}_r \text{ is degenerated})\\
    &= \E[\sup_{r\in R}|\sum_{\pair{i,j}\in E} w_{i,j}(\tilde{h}_r(X_i,X_j^\prime,\tilde{Y}_{i,j}^{\prime\prime})\\
    &\qquad-\E_{\Sigma^{\prime\prime}}[\tilde{h}_r(X_i,X_j^{\prime\prime},\tilde{Y}_{i,j}^{\prime\prime})])|^q]\\
    &(\text{Jansen's Inequality})\\
    &\le \E[\sup_{r\in R}|\sum_{\pair{i,j}\in E} w_{i,j}(\tilde{h}_r(X_i,X_j^\prime,\tilde{Y}_{i,j}^{\prime\prime})\\
    &\qquad-\tilde{h}_r(X_i,X_j^{\prime\prime},\tilde{Y}_{i,j}^{\prime\prime}))|^q]\\
    &= \E[\sup_{r\in R}|\sum_{j=1}^n \sum_{i:\pair{i,j}\in E} w_{i,j}(\tilde{h}_r(X_i,X_j^\prime,\tilde{Y}_{i,j}^{\prime\prime})\\
    &\qquad-\tilde{h}_r(X_i,X_j^{\prime\prime},\tilde{Y}_{i,j}^{\prime\prime}))|^q]
\end{align*}
\DoubleColumnEnd
    \SingleColumn
\begin{align*}
    \E[&\sup_{r\in R}|\sum_{\pair{i,j}\in E} w_{i,j}\tilde{h}_r(X_i,X_j^\prime,\tilde{Y}_{i,j}^{\prime\prime})|^q]\\
    &(\tilde{h}_r \text{ is degenerated})\\
    &= \E[\sup_{r\in R}|\sum_{\pair{i,j}\in E} w_{i,j}(\tilde{h}_r(X_i,X_j^\prime,\tilde{Y}_{i,j}^{\prime\prime})-\E_{\Sigma^{\prime\prime}}[\tilde{h}_r(X_i,X_j^{\prime\prime},\tilde{Y}_{i,j}^{\prime\prime})])|^q]\\
    &(\text{Jansen's Inequality})\\
    &\le \E[\sup_{r\in R}|\sum_{\pair{i,j}\in E} w_{i,j}(\tilde{h}_r(X_i,X_j^\prime,\tilde{Y}_{i,j}^{\prime\prime})-\tilde{h}_r(X_i,X_j^{\prime\prime},\tilde{Y}_{i,j}^{\prime\prime}))|^q]\\
    &= \E[\sup_{r\in R}|\sum_{j=1}^n \sum_{i:\pair{i,j}\in E} w_{i,j}(\tilde{h}_r(X_i,X_j^\prime,\tilde{Y}_{i,j}^{\prime\prime})-\tilde{h}_r(X_i,X_j^{\prime\prime},\tilde{Y}_{i,j}^{\prime\prime}))|^q]
\end{align*}
\SingleColumnEnd
Let $(\sigma_i)_{i=1}^n$ be independent Rademacher variables, independent of $\Sigma$, $\Sigma^\prime$ and $\Sigma^{\prime\prime}$, then we have:
\DoubleColumn
\begin{align*}
    \E[&\sup_{r\in R}|\sum_{j=1}^n \sum_{i:\pair{i,j}\in E} w_{i,j}(\tilde{h}_r(X_i,X_j^\prime,\tilde{Y}_{i,j})\\
    &\qquad-\tilde{h}_r(X_i,X_j^{\prime\prime},\tilde{Y}_{i,j}^{\prime\prime}))|^q\mid \Sigma]\\
    &= \E[\sup_{r\in R}|\sum_{j=1}^n \sigma_j \sum_{i:\pair{i,j}\in E} w_{i,j}(\tilde{h}_r(X_i,X_j^\prime,\tilde{Y}_{i,j})\\
    &\qquad-\tilde{h}_r(X_i,X_j^{\prime\prime},\tilde{Y}_{i,j}^{\prime\prime}))|^q\mid \Sigma]\\
    &(\text{Triangle's Inequality and the convexity of }\sup_{r\in R}|\cdot|^q)\\
    &\le 2^q\E[\sup_{r\in R}|\sum_{j=1}^n \sigma_j \sum_{i:\pair{i,j}\in E} w_{i,j}\tilde{h}_r(X_i,X_j^\prime,\tilde{Y}_{i,j})|^q\mid \Sigma]\\
    &+ 2^q\E[\sup_{r\in R}|\sum_{j=1}^n \sigma_j \sum_{i:\pair{i,j}\in E} w_{i,j}\tilde{h}_r(X_i,X_j^{\prime\prime},\tilde{Y}_{i,j}^{\prime\prime}))|^q\mid \Sigma]\\
    &\le 2^q\E[\sup_{r\in R}|\sum_{j=1}^n\sigma_j\sum_{i:\pair{i,j}\in E}w_{i,j} \tilde{h}_r(X_i,X_j^\prime,\tilde{Y}_{i,j})|^q\mid \Sigma]
\end{align*}
\DoubleColumnEnd
\SingleColumn
\begin{align*}
    \E[&\sup_{r\in R}|\sum_{j=1}^n \sum_{i:\pair{i,j}\in E} w_{i,j}(\tilde{h}_r(X_i,X_j^\prime,\tilde{Y}_{i,j})-\tilde{h}_r(X_i,X_j^{\prime\prime},\tilde{Y}_{i,j}^{\prime\prime}))|^q\mid \Sigma]\\
    &= \E[\sup_{r\in R}|\sum_{j=1}^n \sigma_j \sum_{i:\pair{i,j}\in E} w_{i,j}(\tilde{h}_r(X_i,X_j^\prime,\tilde{Y}_{i,j})-\tilde{h}_r(X_i,X_j^{\prime\prime},\tilde{Y}_{i,j}^{\prime\prime}))|^q\mid \Sigma]\\
    &(\text{Triangle's Inequality and the convexity of }\sup_{r\in R}|\cdot|^q)\\
    &\le 2^q\E[\sup_{r\in R}|\sum_{j=1}^n \sigma_j \sum_{i:\pair{i,j}\in E} w_{i,j}\tilde{h}_r(X_i,X_j^\prime,\tilde{Y}_{i,j})|^q\mid \Sigma]\\
    &\qquad+ 2^q\E[\sup_{r\in R}|\sum_{j=1}^n \sigma_j \sum_{i:\pair{i,j}\in E} w_{i,j}\tilde{h}_r(X_i,X_j^{\prime\prime},\tilde{Y}_{i,j}^{\prime\prime}))|^q\mid \Sigma]\\
    &\le 2^q\E[\sup_{r\in R}|\sum_{j=1}^n\sigma_j\sum_{i:\pair{i,j}\in E}w_{i,j} \tilde{h}_r(X_i,X_j^\prime,\tilde{Y}_{i,j})|^q\mid \Sigma]
\end{align*}
\SingleColumnEnd
and get
\DoubleColumn
\begin{align*}
    \E[\sup_{r\in R}&|\sum_{\pair{i,j}\in E} w_{i,j}\tilde{h}_r(X_i,X_j^\prime,\tilde{Y}_{i,j})|^q] \\
    &\le 2^q\E[\sup_{r\in R}|\sum_{j=1}^n\sigma_j\sum_{i:\pair{i,j}\in E}w_{i,j} \tilde{h}_r(X_i,X_j^\prime,\tilde{Y}_{i,j})|^q]
\end{align*}
\DoubleColumnEnd
\SingleColumn
\begin{align*}
    \E[\sup_{r\in R}|\sum_{\pair{i,j}\in E} w_{i,j}\tilde{h}_r(X_i,X_j^\prime,\tilde{Y}_{i,j})|^q] \le 2^q\E[\sup_{r\in R}|\sum_{j=1}^n\sigma_j\sum_{i:\pair{i,j}\in E}w_{i,j} \tilde{h}_r(X_i,X_j^\prime,\tilde{Y}_{i,j})|^q]
\end{align*}
\SingleColumnEnd
Then repeating the same argument but for the $(X_i)_{i=1}^n$ will give the similar inequality. The desired inequality will follows putting these two inequalities together.
\end{proof}

\begin{proof}[Proof of Theorem~\ref{th:moment_inequality_of_weighted_u_statistics}] 
By the decoupling, undecoupling and randomization techniques (see Lemma~\ref{le:weighted_u_statistics_decoupling}, Lemma~\ref{le:weighted_u_statistics_randomization}), the symmetry and the degeneration of $f$ and the symmetry of $(w_{i,j})_{\pair{i,j}\in E}$, we have
\begin{align*}
    \E[\sup_f&|\sum_{\pair{i,j}\in E}w_{i,j} f(X_i,X_j)|^q]\\
    &\le 16^q\E[\sup_f|\sum_{\pair{i,j}\in E}w_{i,j}\rademacher_i\rademacher_j^\prime f(X_i, X_j^\prime)|^q]\\
    &\le 64^q\E[\sup_f|\sum_{\pair{i,j}\in E}w_{i,j}\rademacher_i\rademacher_j f(X_i, X_j)|^q]
\end{align*}
It means we can convert the moment of the original $U$-process to the moment of Rademacher chaos which can be handled by moment inequalities of \cite{Boucheron2005}.

In particular, for any $q\ge 2$,
\begin{align*}
    (\E_\rademacher[Z_\rademacher^q])^{1/q}
    &\le \E_\rademacher[Z_\rademacher] + (E_\rademacher[(Z_\rademacher-\E_\rademacher[Z_\rademacher])_+^q])^{1/q}\\
    &\le \E_\rademacher[Z_\rademacher]+3\sqrt{q}\E_\rademacher U_\rademacher + 4qB
\end{align*}
where $B$ is defined below
\[B=\sup_f\sup_{\alpha,\alpha^\prime:\|\alpha\|_2,\|\alpha^\prime\|_2\le 1}|\sum_{\pair{i,j}\in E}w_{i,j}\alpha_i\alpha_j^\prime f(X_i,X_j)|.\]
The second inequality above follows by Theorem~$14$ of \cite{Boucheron2005}.

Using the inequality $(a+b+c)^q\le 3^{(q-1)}(a^q+b^q+c^q)$ valid for $q\ge 2,a,b,c>0$, we have
\[\E_\rademacher[Z_\rademacher^q]\le 3^{q-1}(\E_\rademacher[Z_\rademacher]^q+3^qq^{q/2}\E_\rademacher[U_\rademacher]^q+4^qq^qB^q).\]
It remains to derive suitable upper bounds for the expectation of the three terms on the right hand side.

\textit{First term:} $\E[\E_\rademacher[Z_\rademacher]^q]$. Using the symmetrization trick, we have
\[\E[\E_\rademacher[Z_\rademacher]^q]\le 4^q\E[\E_\rademacher[Z_\rademacher^\prime]^q]\]
which $Z_\rademacher^\prime = \sup_f|\sum_{\pair{i,j}\in E}\rademacher_i\rademacher_j^\prime f(X_i,X_j)|$. Note that $\E_\rademacher$ now denotes expectation taken with respect to both the $\rademacher$ and the $\rademacher^\prime$. For simplicity, we denote by $A=\E_{\rademacher}[Z_\rademacher^\prime]$. In order to apply Corollary~$3$ of \cite{Boucheron2005}, define, for $k=1,\dots,n$, the random variables
\[A_k = \E_{\rademacher}[\sup_f |\sum_{\pair{i,j}\in E,i,j\neq k}w_{i,j}\rademacher_i\rademacher_j^\prime f(X_i,X_j)|].\]
It is easy to see that $A_k\le A$.

On the other hand, defining
\[R_k=\sup_f|\sum_{i:\pair{i,k}\in E}w_{i,k}\rademacher_if(X_i,X_k)|,\]
\[M=\max_kR_k\]
and denoting by $f^*$ the function achieving the maximum in the definition of $Z$, we clearly have
\begin{align*}
    A-A_k
    &\le 2\E_\rademacher[M]
\end{align*}
and
\begin{align*}
    \sum_{k=1}^n(A-A_k)
    &\le 2A.
\end{align*}
Therefore,
\[\sum_{k=1}^n(A-A_k)^2\le 4A\E_\rademacher M.\]
Then by Corollary~$3$ of \cite{Boucheron2005}, we obtain
\begin{equation}
    \label{eq:inequality_rademacher_chaos_1}
    \E[\E_\rademacher[Z_\rademacher^\prime]^q]\le 2^{q-1}(2^q(E[Z_\rademacher^\prime])^q+5^qq^q\E[\E_\rademacher[M]^q])
\end{equation}

To bound $\E[\E_\rademacher[M]^q]$, observe that $\E_\rademacher[M]$ is a conditional Rademacher average, for which Theorem~$13$ of \cite{Boucheron2005} could be applied. Since $\max_k\sup_{f,i}w_{i,k}f(X_i,X_k)\le \|\weight{}\|_\infty F$, we have
\begin{equation}
    \label{eq:inequality_rademacher_chaos_2}
    \E[\E_\rademacher[M]^q]\le 2^{q-1}(2^q\E[M]^q+5^qq^q\|\weight{}\|_\infty^qF^q).
\end{equation}
By undecoupling, we have $\E[Z_\rademacher^\prime]=\E[\E_{\rademacher,\rademacher^\prime}[Z_\rademacher^\prime]]\le\E[4\E_\rademacher[Z_\rademacher]]=4\E[Z_\rademacher]$. Collecting all terms, we have
\DoubleColumn
\begin{align*}
    \E[\E_\rademacher[Z_\rademacher]^q]&\le 64^q\E[Z_\rademacher]^q+160^qq^q\E[M]^q\\
    &+400^q\|\weight{}\|_\infty^qF^qq^{2q}.
\end{align*}
\DoubleColumnEnd
\SingleColumn
\begin{align*}
    \E[\E_\rademacher[Z_\rademacher]^q]\le 64^q\E[Z_\rademacher]^q+160^qq^q\E[M]^q+400^q\|\weight{}\|_\infty^qF^qq^{2q}.
\end{align*}
\SingleColumnEnd

\textit{Second term:} $\E[\E_\rademacher[U_\rademacher]^q].$

By the Cauchy-Schwartz inequality we can observe that
\[\sup_{f,i}\sup_{\alpha:\|\alpha\|_2\le 1}\sum_{j:\pair{i,j}\in E}w_{i,j}\alpha_jf(X_i,X_j)\le \|\weight{}\|_{\max}F.\]
Then similar to the bound of $\E[\E_\rademacher[M]^q]$, we have
\[\E[\E_\rademacher[U_\rademacher]^q]\le 2^{q-1}(2^q\E[U_\rademacher]^q+5^qq^q\|\weight{}\|_{\max}^qF^q).\]

\textit{Third term:} $\E[B^q]$. By the Cauchy-Schwartz inequality, we have $B\le \sqrt{\sum_{\pair{i,j}\in E}w_{i,j}^2}F= \|\weight{}\|_2F$ so
\[\E[B^q]\le \|\weight{}\|_2^qF^q.\]

Now it remains to simply put the pieces together to obtain
\DoubleColumn
\begin{align*}
\E[\sup_f&|\sum_{\pair{i,j}\in E}w_{i,j} f(X_i,X_j)|^q]\\
&\le C(\E[Z_\rademacher]^q+q^{q/2}\E[U_\rademacher]^q+q^q\E[M]^q\\
&\qquad+\|\weight{}\|_\infty^q F^qq^{2q}+\|\weight{}\|_{\max}^qF^qq^{3q/2}\\
&\qquad+F^q\|\weight{}\|_2^qq^q)
\end{align*}
\DoubleColumnEnd
\SingleColumn
\begin{align*}
\E[\sup_f&|\sum_{\pair{i,j}\in E}w_{i,j} f(X_i,X_j)|^q]\\
&\le C(\E[Z_\rademacher]^q+q^{q/2}\E[U_\rademacher]^q+q^q\E[M]^q+\|\weight{}\|_\infty^q F^qq^{2q}+\|\weight{}\|_{\max}^qF^qq^{3q/2}+F^q\|\weight{}\|_2^qq^q)
\end{align*}
\SingleColumnEnd
for an appropriate constant $C$. 
In order to derive the exponential inequality, we use Markov inequality $\Pro[X\ge t]\le t^{-q}\E[Z^q]$ and choose
\DoubleColumn
\begin{align*}
    q&=C\min\Big((\frac{t}{\E[U_\rademacher]})^2,\frac{t}{\E[M]},\frac{t}{F\|\weight{}\|_2},(\frac{t}{\|\weight{}\|_{\max} F})^{2/3},\\
    &\qquad\sqrt{\frac{t}{\|\weight{}\|_\infty F}}\Big)
\end{align*}
\DoubleColumnEnd
\SingleColumn
\begin{align*}
    q=C\min\Big((\frac{t}{\E[U_\rademacher]})^2,\frac{t}{\E[M]},\frac{t}{F\|\weight{}\|_2},(\frac{t}{\|\weight{}\|_{\max} F})^{2/3},\sqrt{\frac{t}{\|\weight{}\|_\infty F}}\Big)
\end{align*}
\SingleColumnEnd
for an appropriate constant $C$.
\end{proof}

\begin{proof}[Proof of Corollary~\ref{cor:covering_number_inequality}] 
    From Theorem~$3$, it is easy to know
    \DoubleColumn
    \begin{equation}
        \begin{aligned}
            \label{eq:moment_inequality_of_weighted_u_statistics}
            \kappa=&C(\E[Z_\rademacher]+\max(\E[U_\rademacher]\sqrt{\log(1/\delta)},\E[M]\log(1/\delta),\\
            &\quad\log(1/\delta)\|\weight{}\|_2,(\log(1/\delta))^{3/2}\|\weight{}\|_{\max},\\
            &\quad (\log(1/\delta))^2\|\weight{}\|_\infty)).
        \end{aligned}
    \end{equation}
    \DoubleColumnEnd
    \SingleColumn
    \begin{equation}
        \begin{aligned}
            \label{eq:moment_inequality_of_weighted_u_statistics}
            \kappa=C(\E[Z_\rademacher]+\max(&\E[U_\rademacher]\sqrt{\log(1/\delta)},\E[M]\log(1/\delta),\log(1/\delta)\|\weight{}\|_2,(\log(1/\delta))^{3/2}\|\weight{}\|_{\max},\\
            &\quad (\log(1/\delta))^2\|\weight{}\|_\infty)).
        \end{aligned}
    \end{equation}
    \SingleColumnEnd
An important character of these Rademacher processes in \eqref{eq:moment_inequality_of_weighted_u_statistics} is that they all satisfy the Khinchine inequality \eqref{eq:kinchine_type_inequality_process}. For simplicity, we denote the weighted Rademacher processes of $Z_\rademacher$ by 
\[\{z_\rademacher(f)=\sum_{\pair{i,j}\in E_<}w_{i,j}\rademacher_i\rademacher_j\linebreak f(X_i,X_j), f\in\mathcal{F}\}.\]
where $E_<=\{(i,j): \{i,j\}\in E, i<j\}$. Let $z_\rademacher^\prime=z_\rademacher/\|\weight{}\|_2$, following Theorem \ref{th:kinchine_type_ineqaulity}, we can easily have that $z_\rademacher^\prime$ satisfies \eqref{eq:kinchine_type_inequality_process} with degree $2$. Thus from Theorem \ref{th:rademacher_process_chaining}, we have
\begin{equation}
    \label{eq:weighted_rademacher_process_prove_1}
    \E_\rademacher[\sup_{f,g}|z^\prime_\rademacher(f)-z^\prime_\rademacher(s)|]\le K\int_0^D\log N(\mathcal F,\lnorm{}_2,\epsilon)d\epsilon.
\end{equation}
Recall that $\sup_f \|f\|_\infty = F$, we have $D= 2F$.
Since this metric function $d$ is intractable, we need to convert it to the $\lnorm_\infty$ metric. For all $f,s$, we have
\DoubleColumn
$\lnorm{}_2(z'_\rademacher{}(f),z'_\rademacher{}(g))\le \lnorm_\infty(f,g)$.
\DoubleColumnEnd
\SingleColumn
\begin{align*}
    \lnorm{}_2(f,g)&\le \lnorm_\infty(f,g)
\end{align*}
\SingleColumnEnd
The fact that if $\forall f,s\in\mathcal F, \lnorm{}(f,s)\le \lnorm{}'(f,s)$ then $N(\mathcal F,\lnorm{},\epsilon)\le N(\mathcal F, \lnorm{}',\epsilon)$ combined with \eqref{eq:weighted_rademacher_process_prove_1} will give
\begin{align*}
    \E[Z_\rademacher]&= 2\E[\E_\rademacher[\sup_f|z_\rademacher(f)|]]\\
    &\le 2K\|\weight{}\|_2\int_0^{2F}\log N(\mathcal F,\lnorm{}_2,\epsilon)d\epsilon\\
    &\le 2K\|\weight{}\|_2\int_0^{2F}\log N_\infty(\mathcal F, \epsilon)d\epsilon
\end{align*}
where $K$ is a universal constant. Similarly, we can also bound $\E[M]$ by $K\|\weight{}\|_{\max}\linebreak\int_0^{2F}\sqrt{\log N_\infty(\mathcal F, \epsilon)}d\epsilon$. 
For $U_\rademacher$, let $\alpha^*$ be the (random) vector that maximizes $U_\rademacher$ and define
\[\{u_\rademacher(f) = \sum_{i=1}^n\rademacher_i\sum_{\pair{i,j}\in\E}w_{i,j}\alpha_j^*\linebreak f(X_i,X_j),f\in\mathcal{F}\}.\]
Clearly, $u_\rademacher$ satisfies the Khintchine inequality with degree $1$. Also, we need to convert its metric distance and $\lnorm{}_2(u_\rademacher{}(f), u_\rademacher{}(g))\le \lnorm{}_\infty(f, g)$.
Thus, $\E[U_\rademacher]\le K\|\weight{}\|_2\int_0^{2F}\sqrt{\log N_\infty(\mathcal F,\epsilon)}d\epsilon$.

Plugging all these part into \eqref{eq:moment_inequality_of_weighted_u_statistics} will complete the corollary.
\end{proof}

\section{Metric Entropy Inequality} 
\label{sub:metric_entropy_inequality}

The following theorems are more or less classical and well known. We present them here for the sake of completeness.

\begin{theorem}[Khinchine inequality for Rademacher chaos, {\citett{Theorem 3.2.1}{de2012decoupling}}]
    \label{th:kinchine_type_ineqaulity}
    Let $F$ be a normed vector space and let $\{\rademacher{}_i\}_{i=1}^\infty$ be a Rademacher sequence. Denote by
    \DoubleColumn
    \begin{align*}
        X=x+\sum_{i=1}^n x_i\rademacher_i+\sum_{i_1<i_2\le n}\linebreak x_{i_1,i_2}\rademacher{}_{i_1}\rademacher{}_{i_2}+\dots\\
        +\sum_{i_1<\dots<i_d\le n}x_{i_1\dots i_d}\rademacher{}_{i_1}\dots \rademacher{}_{i_d}        
    \end{align*}
    \DoubleColumnEnd
    \SingleColumn
    \begin{align*}
        X=x+\sum_{i=1}^n x_i\rademacher_i+\sum_{i_1<i_2\le n}\linebreak x_{i_1,i_2}\rademacher{}_{i_1}\rademacher{}_{i_2}+\dots+\sum_{i_1<\dots<i_d\le n}x_{i_1\dots i_d}\rademacher{}_{i_1}\dots \rademacher{}_{i_d}        
    \end{align*}
    \SingleColumnEnd
    the Rademacher chaos of order d. Let $1<p\le q<\infty$ and let
    \[\gamma=(\frac{p-1}{q-1})^{1/2}.\]
    Then, for all $d\ge 1$,
    \DoubleColumn
    \begin{align*}
        (\E[|x&+\sum_{i=1}^n \gamma x_i\rademacher_i+\sum_{i_1<i_2\le n}\gamma^2x_{i_1,i_2}\rademacher{}_{i_1}\rademacher{}_{i_2}+\dots\\
        &\qquad+\sum_{i_1<\dots<i_d\le n}\gamma^dx_{i_1\dots i_d}\rademacher{}_{i_1}\dots \rademacher{}_{i_d}|^q])^{1/q}\\
        &\le (\E[|x+\sum_{i=1}^n x_i\rademacher_i+\sum_{i_1<i_2\le n}x_{i_1,i_2}\rademacher{}_{i_1}\rademacher{}_{i_2}+\dots\\
        &\qquad+\sum_{i_1<\dots<i_d\le n}x_{i_1\dots i_d}\rademacher{}_{i_1}\dots \rademacher{}_{i_d}|^p])^{1/p}
    \end{align*}
    \DoubleColumnEnd
    \SingleColumn
    \begin{align*}
        (\E[|x&+\sum_{i=1}^n \gamma x_i\rademacher_i+\sum_{i_1<i_2\le n}\gamma^2x_{i_1,i_2}\rademacher{}_{i_1}\rademacher{}_{i_2}+\dots+\sum_{i_1<\dots<i_d\le n}\gamma^dx_{i_1\dots i_d}\rademacher{}_{i_1}\dots \rademacher{}_{i_d}|^q])^{1/q}\\
        &\le (\E[|x+\sum_{i=1}^n x_i\rademacher_i+\sum_{i_1<i_2\le n}x_{i_1,i_2}\rademacher{}_{i_1}\rademacher{}_{i_2}+\dots+\sum_{i_1<\dots<i_d\le n}x_{i_1\dots i_d}\rademacher{}_{i_1}\dots \rademacher{}_{i_d}|^p])^{1/p}
    \end{align*}
    \SingleColumnEnd
\end{theorem}

\begin{theorem}[metric entropy inequality, {\citett{Proposition 2.6}{Dembo1994}}]
    \label{th:rademacher_process_chaining}
    If a process $\{Y_f: f\in\mathcal F\}$ satisfies
\begin{equation}
    \label{eq:kinchine_type_inequality_process}
    (\E[|Y_f-Y_g|^p])^{1/p}\le (\frac{p-1}{q-1})^{m/2}(\E[|Y_f-Y_g|^q])^{1/q},
\end{equation}
for $1<q<p<\infty$ and some $m\ge 1$, and if
\begin{equation}
    \label{eq:process_distance}
    d(f,g) = (\E[|Y_f-Y_g|^2])^{1/2},
\end{equation}
there is a constant $K<\infty$ such that
    \begin{equation}
        \label{eq:rademacher_process_chaining}
        \E[\sup_{f,g}|Y_f-Y_g|]\le K\int_0^D(\log N(\mathcal F,d,\epsilon))^{m/2}d\epsilon.
    \end{equation}
    where $D$ is the $d$-diameter of $\mathcal{F}$.
\end{theorem}

\end{document}